\documentclass[twoside]{article}

\usepackage[accepted]{aistats2021}

\usepackage{paralist}
\usepackage[utf8]{inputenc} % allow utf-8 input
\usepackage[T1]{fontenc}    % use 8-bit T1 fonts
\usepackage{xr}
\usepackage{hyperref}       % hyperlinks
\usepackage{url}            % simple URL typesetting
\usepackage{booktabs}       % professional-quality tables
\usepackage{amsfonts}       % blackboard math symbols
\usepackage{nicefrac}       % compact symbols for 1/2, etc.
\usepackage{microtype}      % microtypography

\usepackage[english]{babel}

\usepackage{filecontents}
\usepackage{amsmath}
\usepackage{amssymb}
\usepackage{float}
\usepackage{placeins}
\usepackage{graphicx}
\usepackage{subcaption} % subfigure
\usepackage[dvipsnames]{xcolor}
\usepackage{tikz,tkz-euclide,pgfplots}

\usepackage[numbers]{natbib}

\newcommand{\ie}{i\/.\/e\/.,\/~}
\newcommand{\eg}{e\/.\/g\/.,\/~}
\newcommand{\cf}{cf\/.\/~}

\newcommand*{\parencite}{\citep}
\begin{document}

%
%\twocolumn[
%\aistatstitle{Structure-preserving Gaussian Process Dynamics}
%

%
%\aistatsaddress{Bosch Center for \\ Artificial Intelligence \\ Renningen, Germany  \And DSME, RWTH Aachen University \\ Aachen, Germany \And Bosch Center for \\ Artificial Intelligence \\ Renningen, Germany  }
%
%\aistatsauthor{Michael Tiemann \And Sebastian Trimpe}
%
%\aistatsaddress{ Bosch Center for Artificial Intelligence \\ Renningen, Germany  \And  DSME, RWTH Aachen University \\Aachen, Germany}
%]

\runningauthor{Katharina Ensinger, Friedrich Solowjow, Sebastian Ziesche, Michael Tiemann, Sebastian Trimpe}

\twocolumn[

\aistatstitle{Structure-preserving Gaussian Process Dynamics}
 
\aistatsauthor{%
	\textbf{Katharina Ensinger}$^{1,2}$\hspace{0.4cm} \textbf{Friedrich Solowjow}$^{2}$\hspace{0.4cm}
	  \textbf{Sebastian Ziesche}$^{1}$
	\\
	\textbf{Michael Tiemann}$^{1}$\hspace{0.4cm}
	\textbf{Sebastian Trimpe}$^{2}$\hspace{0.4cm}\vspace{0.2cm}
}

\aistatsaddress{$^{1}$Bosch Center for Artificial Intelligence, Renningen, Germany \hspace{0.4cm} \\$^{2}$ Institute for Data Science in Mechanical Engineering, RWTH Aachen University, Aachen, Germany}

]

\begin{abstract}
	Most physical processes posses structural properties such as constant energies, volumes, and other invariants over time.
	When learning models of such dynamical systems, it is critical to respect these invariants to ensure accurate predictions and physically meaningful behavior.
	Strikingly, state-of-the-art methods in Gaussian process (GP) dynamics model learning are not addressing this issue. On the other hand, classical numerical integrators are specifically designed to preserve these crucial properties through time.
	We propose to combine the advantages of GPs as function approximators with structure preserving numerical integrators for dynamical systems, such as Runge-Kutta methods. These integrators assume access to the ground truth dynamics and require evaluations of intermediate and future time steps that are unknown in a learning-based scenario. This makes direct inference of the GP dynamics, with embedded numerical scheme, intractable.
	Our key technical contribution is the evaluation of the implicitly defined Runge-Kutta transition probability.
	In a nutshell, we introduce an implicit layer for GP regression, which is embedded into a variational inference-based model learning scheme.
\end{abstract}

\section{Introduction} \label{section:intro}
Many physical processes can be described by an autonomous continuous-time dynamical system
\begin{equation}\label{eq:dyn}
\dot{x}(t)=f(x(t)) \textrm{ with } f:\mathbb{R}^d \rightarrow \mathbb{R}^d.
\end{equation}
Dynamics model learning deals with the problem of estimating the function $f$ from sampled data.
In practice, it is not possible to observe state trajectories in continuous time since data is typically collected on digital sensors and hardware.
Thus, we obtain noisy discrete-time observations
\begin{equation} \label{eq:noisy-data}
\begin{aligned}
\{\hat{x}_n \}_{1:N} & = \{x_{1}+\nu_1, \dots, x_{N}+\nu_N \}, \\
\nu_n  & \sim \mathcal{N}(0,\mathrm{diag}(\sigma_{n,1}^2,\dots,\sigma_{n,d}^2)).
\end{aligned}
\end{equation}
Accordingly, models of dynamical systems are typically learned as of one-step ahead-predictions
\begin{equation}\label{eq:oneStep}
x_{n+1}=g(x_n).
\end{equation}
Especially, Gaussian processes (GPs) have been popular for model learning  and are predominantly applied to one step ahead predictions \eqref{eq:oneStep} \parencite{PILCO, Frigola, doerr2018probabilistic, pmlr-v120-buisson-fenet20a}.
However, there is a discrepancy between the continuous \eqref{eq:dyn} and discrete-time \eqref{eq:oneStep} systems.
Importantly, \eqref{eq:dyn} often posseses invariants that represent physical properties.
Thus, naively chosen discretizations might lead to poor models.

Numerical integrators provide sophisticated tools to efficiently discretize continuous-time dynamics~\eqref{eq:dyn}.
Strikingly, one step ahead predictions \eqref{eq:oneStep} correspond to the explicit Euler integrator $x_{n+1}=x_n+hf(x_n)$ with step size $h$.
This follows immediately by identifying $g(x_n)$ with $x_n+hf(x_n)$.
It is well-known that the explicit Euler method might lead to problematic behavior and suboptimal performance \parencite{hairerVol1}.
Clearly, this raises the immediate question: \emph{can superior numerical integrators be leveraged for dynamics model learning?}
For the numerical integration of dynamical systems, the function $f$ is assumed to be known.
The explicit Euler is a popular and straightforward method, which thrives on its simplicity.
No intermediate evaluations of the dynamics are necessary, which makes the integrator also attractive for model learning. 
While this behavior is tempting when implementing the algorithm, there are theoretical issues \parencite{hairerVol1}.
In particular, important physical and geometrical structure is not preserved.
In contrast to the explicit Euler, there are also implicit and higher-order methods.
However, these generalizations require the evaluation at intermediate and future time steps, which leads to a nonlinear system of equations that needs to be solved.
While these schemes become more involved, they yield advantageous theoretical guarantees.
In particular, Runge-Kutta (RK) schemes define a rich class of powerful integrators.
Despite assuming the dynamics function $f$ to be unknown, we can still benefit from the discretization properties of numerical integrators for model learning.
To this end, we propose to combine GP dynamics learning with arbitrary RK integrators, in particular, implicit RK integrators.

Depending on the problem that is addressed, the specific RK method has to be tailored to the system.
As an example, we consider structure-preserving integrators, \ie geometric and symplectic ones.
We develop our arguments based on Hamiltonian systems  \parencite{ham, Sakurai:1167961, matterPhysics}.
These are an important class of problems that preserve a generalized notion of energy and volume.
Symplectic integrators are designed to cope with this type of problems providing volume-preserving trajectories and accurate approximation of the total energy \parencite{hairer2006geometric}.
In order to demonstrate the flexibility of our method, we also introduce a geometric integrator that is consistent with a mass moving on a surface.
For both examples, we show in the experiments section that the predictions with our tailored GP model are indeed preserving the desired structure.

By generalizing to more sophisticated integrators, we have to address the issue of propagating implicitly defined distributions through the dynamics.
This is due to the fact that evaluations of the GP at the next time step induce additional implicit evaluations of the dynamics.
Depending on the integrator, these might be future or intermediate time steps that are also unknown.
On a technical level, sparse GPs provide the necessary flexibility.
A decoupled sampling approach allows consistent sampling of functions from the GP posterior \parencite{wilson2020efficiently}.
In contrast to previous GP dynamics modeling schemes this yields consistency throughout the entire simulation pipeline \parencite{IalWilHenRas19}. 
By leveraging these ideas, we derive a recurrent variational inference (VI) model learning scheme.

By addressing integrator-induced implicit transition probabilities, we are essentially proposing implicit layers for probabilistic models.
Implicit layers in neural networks (NNs) are becoming increasingly popular \parencite{gould2019deep, equilibrium, look2020differentiable}.
However, the idea of implicitly defined layers has (to the best of our knowledge) not yet been generalized to probabilistic models like GPs.

In summary, the main contributions of this paper are:
\begin{compactitem}
	\item a general and flexible probabilistic learning framework that combines arbitrary RK integrators with GP dynamics model learning;
	\item deriving an inference scheme that is able to cope with implicitly defined distributions, thus extending the idea of implicit layers from NNs to probabilistic GP models; and
	\item embedding geometrical and symplectic integrators, yielding structure-preserving GP models.
\end{compactitem}

\section{Related work}
Dynamics model learning is a very broad field and has been addressed by various communities for decades, e.g., \citep{dynRobot, schon2011system, ljung1999system, geist2020learning}.
Learning dynamics models can be addressed with a continuous time model \citep{heinonen2014learning, Wenketal19}. 
Nonparametric models were learned by applying sparse GPs  \parencite{DBLP:conf/icml/HeinonenYMIL18, hegde2021bayesian}.
A common approach for learning discrete time models are Gaussian process state-space models \parencite{Turner, Wang, Lindholm}.
In this work we consider fully observable models in contrast to common state-space models.
However, we apply the tools of state-space model literature.
We develop our ideas exemplary for the inference scheme proposed in  \citet{doerr2018probabilistic}. At the same time, our contribution is not restricted to that choice of inference scheme and can be combined with other schemes as well. 
\citet{IalWilHenRas19} have criticized the inference scheme and shown that there might be issues in the presence of transition noise and the inference might become inconsistent.
However, none of these concerns are relevant to our work since we omit transition noise and sample GPs from the posterior (\cf Sec. \ref{section:method}.)

Implicit transitions have become popular for NNs and provide useful tools and ideas that we leverage. 
In general, implicit transitions are cast into an optimization problem \parencite{gould2019deep}. 
On a technical level, we implement related techniques based on the implicit function theorem and backpropagation.
In \citet{NEURIPS2020_83eaa672}, implicit transition functions are used to perform multi-modal regression tasks. 
A NN with infinitely many layers was trained by implicitly defining the equilibrium of the network \parencite{equilibrium}.
\citet{look2020differentiable} proposed an efficient backpropagation scheme for implicitly defined neural network layers. 
All these approaches refer to deterministic NNs, while we we extend the ideas to probabilistic GP models. 

Including Hamiltonian structure into learning in order to preserve physical properties of the system, is an important problem addressed by many sides.
The problem can be tackled by approximating the continuous time Hamiltonian structure from data. 
This was addressed by applying a NN \parencite{NIPS20199672}.
Since modern NN approaches provide a challenging benchmark, we compare our method against \citet{NIPS20199672}.
In \citet{jin2020sympnets}, the Hamiltonian structure is learned by stacking multiple symplectic modules.
Gaussian processes have been combined with symplectic structure \citep{rath2020symplectic}.
In contrast to our approach, the focus lies on learning continuous-time dynamics for Hamiltonian systems and unrolling the dynamics via certain symplectic integrators. 
In \citet{2021structurepreserving} variational integrators are applied to a continuous-time GP dynamics model. In contrast to our approach the integrator step is decoupled from the learning step. 

However, there is literature that adresses discrete time Hamiltonian systems. 
The Hamiltonian neural network approach was extended by a recurrent NN coupled with a symplectic integrator \citep{chen2019symplectic}.
In \citet{saemundsson2020variational}, the symplectic approach was combined with uncertainty information by applying variational autoencoders.
\citet{zhong2020symplectic} extends previous approaches by adding control input. 
The stable nature of explicit symplectic integrators is leveraged to build the architecture of a NN by introducing symplectic transitions between the layers of a NN \parencite{Haber}.
 In contrast to previous approaches, we are able to address non-separable Hamiltonians via implicit integrators. 

% !TEX root = aistats_2022.tex
\section{Technical background and main idea}
\label{section:background}
Next, we make our problem mathematically precise and provide a summary of the preliminaries.
\subsection{Gaussian process regression}
\label{section:varGP}
A GP is a distribution over functions \parencite{10.5555/1162254}. %Rasmussen.
Similar to a normal distribution, a GP is determined by its mean function $m(x)$ and covariance function $k(x,y)$. We assume the prior mean to be zero.

\textbf{Standard GP inference:} For direct training, the GP predictive distribution is obtained by conditioning on $n$ observed data points. In addition to optimizing the hyperparameters, a system of equations has to be solved, which has a complexity of $\mathcal{O}(n^3)$. Clearly, this is problematic for large datasets.

\textbf{Variational sparse GP:}
The GP can be sparsified by introducing pseudo inputs \parencite{Titsias}.
Intuitively, we approximate the posterior with a lower number of training points.
However, this makes direct inference intractable.
An elegant approximation strategy is based on casting Bayesian inference as an optimization problem.
We consider pseudo inputs $\xi = [\xi_1, \dots, \xi_P] $ and targets $z = [z_1, \dots, z_P]$ as proposed in \parencite{Hensman} and applied in \parencite{doerr2018probabilistic, IalWilHenRas19, foell2019deep}. Intuitively, the targets can be interpreted as GP observations at $\xi$.
The posterior of pseudo targets is approximated via a variational approximation $q(z)= \mathcal{N}(\mu,\Sigma)$, where $\mu$ and $\Sigma$ are adapted during training.
The GP posterior distribution at inputs $x^{*}$ is conditioned on the pseudo inputs resulting in a normal distribution $f(x^{*}|z,\xi) \sim \mathcal{N}(\mu(x^*) ,\Sigma(x^*))$ with
\begin{equation} \label{eq:sparse-moments}
\begin{aligned}
\mu(x^*) & = k(x^{\star},\xi)k(\xi,\xi)^{-1}z \\
\Sigma(x^{\star})& = k(x^{\star},x^{\star})- k(x^{\star},\xi)k(\xi,\xi)^{-1}k(\xi, x^{\star}).
\end{aligned}
\end{equation}

\textbf{Decoupled sampling: }
In contrast to standard (sparse) GP conditioning this allows to sample globally consistent functions from the posterior.
Thus, succesively sampling at multiple inputs is achieved without conditioning.
By applying Matheron's rule \parencite{howarth1979},
the GP posterior is decomposed into two parts~\parencite{wilson2020efficiently},
\begin{equation}\label{eq:matheron}
\begin{aligned}
f(x^{\star}|z, \xi)&=\underbrace{f(x^{\star})}_{\textrm{prior}}+\underbrace{k(x^{\star},\xi)k(\xi,\xi)^{-1}(z-f_z)}_{\textrm{update}} \\
& \approx \sum_{i=1}^{S} w_i \phi_i(x^{\star})+\sum_{j=1}^M v_j k(x^{\star},\xi_j)(z-f_z),
\end{aligned}
\end{equation}
where $S$ Fourier bases $\phi_i$ and $w_i \sim \mathcal{N}(0,1)$ represent the stationary GP prior \parencite{Rahimi}.
For the update it holds that $v = k(\xi,\xi)^{-1}(z-\Phi W), \Phi = \phi(\xi) \in \mathbb{R}^{S \times D}$.
The targets $z$ are sampled from the variational distribution $q(z)$.
We add technical details in the supplementary material.

\subsection{Runge-Kutta integrators}
\label{section:RK}
A RK integrator $\psi_f$ for a continuous-time dynamical system $f$ \eqref{eq:dyn} is designed to approximate the solution $x(t_n)$ at discrete time steps $t_n$ via $\bar{x}_n$. Hence,
\begin{equation} \label{eq:rk}
\begin{aligned}
\bar{x}_{n+1}&  = \psi_f(\bar{x}_n) = \bar{x}_n +h\sum_{j=1}^s b_j g_j, \\
g_j&= f(\bar{x}_n+ h \sum_{l=1}^s a_{jl} g_l), j = 1,\dots,s,
\end{aligned}
\end{equation}
where $g_j$ are the internal stages and $\bar{x}_0 = x(0)$.
We use the notation $\bar{x}$ to indicate numerical error corrupted states and highlight the subtle difference to ground truth data.
The parameters $a_{jl}, b_j  \in \mathbb{R}$ determine the properties of the method, \eg the stability radius of the method \citep{hairerVol2}, the geometrical properties, or whether it is symplectic \citep{hairer2006geometric}.

\textbf{Implicit integrators: } If $a_{jl} > 0$ for $l\ge j$,
Eq.~\eqref{eq:rk} takes evaluations at time steps into account where the state is not yet known.
Therefore, the solution of a nonlinear system of equations is required.
A prominent example is the implicit Euler scheme $\bar{x}_{n+1} = \bar{x}_n + h f(\bar{x}_{n+1})$.
\subsection{Main idea}
We propose to embed RK methods \eqref{eq:rk} into GP regression.
Since the underlying ground truth dynamics $f$ \eqref{eq:dyn} are given in continuous time, the discretization matters.
Naive methods, such as the explicit Euler method, are known to be inconsistent with physical behavior.
Therefore, we investigate how to learn more sophisticated models that, by design, are able to preserve physical structure of the original system.
Further, we will develop this idea into a tractable inference scheme.
In a nutshell, we learn GP dynamics $\hat{f}$ that yield predictions $x_{n+1} = \psi_{\hat{f}}(x_n)$.
This enforces the RK \eqref{eq:rk} instead of explicit Euler \eqref{eq:oneStep} structure.
Thus, leading to properties like volume preservation.
The main technical difficulty lies in making the implicitly defined transition probability $p(\psi_{\hat{f}}(x_n) | x_n)$ tractable.

% !TEX root=aistats_2022.tex
\section{Embbedding Runge-Kutta integrators in GP models}
\label{section:method}
Next, we dive into the technical details of merging GPs with RK integrators.
We demonstrate how to evaluate the implicitly defined transition probability of any higher-order RK method by applying decoupled sampling \eqref{eq:matheron}.
This technique is then applied to a recurrent variational inference scheme in order to obtain a GP dynamics model.
\subsection{Efficient evaluation of the transition model}  \label{section:reparam}
At its core, we consider the problem of evaluating implicitly defined distributions of RK integrators $\psi_{\hat{f}}$.
To this end, we derive a sampling-based technique by leveraging decoupled sampling \eqref{eq:matheron}.
This enables us to perform the integration step on a sample of the GP dynamics $\hat{f}$.
The procedure is illustrated in Figure \ref{subfig:scheme}.
We model the dynamics $\hat{f}$ via $d$ variational sparse GPs.
Let  $z \in \mathrm{R}^{d\times P}$ be a sample from the variational posterior $q(z)$ (\cf Sec. \ref{section:varGP}).
The probabilty of an integrator step $p(\psi_{\hat{f}}(x_n)|z, x_n)$ is formally obtained by integrating over all possible GP dynamics $\hat{f}|z$,
\begin{equation}\label{eq:int}
\begin{aligned}
p(x_{n+1}|z,x_n)&=p(\psi_{\hat{f}}(x_n)|z, x_n) \\
&=\int_{\hat{f}}p(\psi(x_n)|\hat{f})p(\hat{f}|z) d\hat{f}.
\end{aligned}
\end{equation}
Performing an RK integrator step $\psi_{\hat{f}}(x_n)$ requires the computation of RK stages $g^{\star} = (g^{\star}_1, \dots, g^{\star}_s)$ \eqref{eq:rk}
\begin{equation}\label{eq:rep}
g^{\star} = \arg \min_g \Vert g-\hat{f}(x_n+hAg) \Vert^2,
\end{equation}
with $A = (a_{jl})_{j=1,\dots s, l = 1 \dots j}$ determined by the RK scheme \eqref{eq:rk}.
In the explicit case $A$ is a sub-diagonal matrix so  $g_j$ can be calculated successively.
In the implicit case, a minimization problem has to be solved.
In order to sample an integrator step from \eqref{eq:int}, we first compute a GP dynamics function $\hat{f}|z$ via \eqref{eq:matheron}.
This is achieved by sampling inducing targets $z$ from $\mathcal{N}(\mu,\Sigma)$ and $w\sim\mathcal{N}(0,1)$ (\cf Sec \ref{section:varGP}).
We are now able to evaluate $\hat{f}$ at arbitrary locations and perform an RK integrator step with respect to the dynamics $\hat{f}$.
To this end, the system of equations \eqref{eq:rep} is defined and solved with respect to the fixed GP dynamics $\hat{f}$.
This enables us to extend implicit layers to probabilistic models.
Combining \eqref{eq:rep} and \eqref{eq:matheron} with $u = u(g) = x_n + h A g $ and $\hat{z} = z-f_z$ yields
\begin{equation}\label{eq:fullEQ}
\begin{aligned}
g^{\star}= \arg \min_g\Big\Vert\,g-\sum_{i=1}^{S} w_i \phi_i(u)-\sum_{j=1}^M v_j k(u,\xi_j)\hat{z}\,\Big\Vert^2.
\end{aligned}
\end{equation}
%\michael{Need to replace $x^*$ with $x_n + h A g$ or put some definitions in, if the line gets too long otherwise}
Next, we give an example. Consider the IA-Radau method \citep{hairerVol2} $x_{n+1} = x_n+h \left(\frac{1}{4} g_1 + \frac{3}{4} g_2 \right)$, with
\tikzset{every picture/.style={line width=0.75pt}} %set default line width to 0.75pt
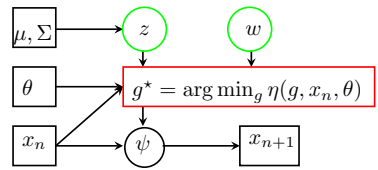
\begin{figure}
	% \begin{subfigure}  [t]{0.48\textwidth}
	% 	\centering
	% 	\input{fig_1.tex}
	% 	\caption{Recurrent scheme}\label{subfig:rec}
	% \end{subfigure}
	% \begin{subfigure}[t]{0.48\textwidth}
		\centering
		\resizebox{0.3\textwidth}{!}{
		\begin{tikzpicture}[x=0.75pt,y=0.75pt,yscale=-1,xscale=1]
\tikzset{every picture/.style={line width=0.75pt}} %set default line width to 0.75pt        


%Shape: Rectangle [id:dp6171132874626408] 
%\x_n

\draw   (20,82) -- (46,82) -- (46,58) -- (20,58) -- cycle ;

%Shape: Rectangle [id:dp5991929739602408] 

%\k^{\star}
%\draw   (87,58) -- (111,58) -- (111,82) -- (87,82) -- cycle ;
%Shape: Rectangle [id:dp8832167115767708] 

%\hat{x}_n+1
\draw   (158,58) -- (194,58) -- (194,82) -- (158,82) -- cycle ;
%Shape: Rectangle [id:dp05162269133020936] 

%\theta
\draw   (20,22) -- (46,22) -- (46,46) -- (20,46) -- (20,22) -- cycle ;

\draw   (20,-14) -- (46,-14) -- (46,10) -- (20,10) -- (20,-14) -- cycle ;

\draw  [->,>=stealth]  (46,70) -> (87,70) ;
\draw  [->,>=stealth]  (111,70) -> (158,70) ;
\draw  [->,>=stealth]  (46,34) -> (87,34) ;
\draw  [->,>=stealth]  (46, 70) -> (87,34) ;
\draw  [->,>=stealth]  (99, 46) -> (99,58) ;
\draw  [->,>=stealth]  (99, 12) -> (99,22) ;
\draw  [->,>=stealth]  (46, -1) -> (87,-1) ;
\draw  [->,>=stealth]  (165, 12) -> (165,22) ;

%Shape: Rectangle [id:dp9045006747352238] 

\draw (100,71) circle(9pt);

\draw[draw = green] (100,-1) circle(10pt); 
\draw[draw = green] (164,-1) circle(10pt); 
%k^{star}
\draw[draw = red]   (87,22) -- (240,22) -- (240,46) -- (87,46) -- cycle ;
%Shape: Ellipse [id:dp7142738141362306] 

%Shape: Rectangle [id:dp8752493862362398] 

% Text Node
\draw (24,28.4) node [anchor=north west][inner sep=0.75pt]    {$\theta$};
% Text Node
\draw (91,28.4) node [anchor=north west][inner sep=0.75pt]    {$g^{\star}=\arg\min_g \eta(g,x_n,\theta)$};
\draw (95,-5) node [anchor=north west][inner sep=0.75pt]    {$z$};
\draw (160,-5) node [anchor=north west][inner sep=0.75pt]    {$w$};
% Text Node
\draw (18.8, -5) node [anchor=north west][inner sep=0.75pt]    {$\mu, \Sigma$};
% Text Node
 
% Text Node

% Text Node
\draw (24,63) node [anchor=north west][inner sep=0.75pt]    {$x_n$};
% Text Node
\draw (93,61.73) node [anchor=north west][inner sep=0.75pt]    {$\psi$};
% Text Node
\draw (162,61.73) node [anchor=north west][inner sep=0.75pt]    {$x_{n+1}$};
 
\end{tikzpicture} 
	}
		% \caption{Integration step}\label{subfig:scheme}
	% \end{subfigure}
	\caption{
	The evaluation of an integrator step.
	%  is illustrated in \ref{subfig:scheme}.
	First, weights $w$ and inducing targets $z$ are sampled. This yields tractable solutions to the minimization problem for the RK stages $g^{\star}$ (red).
	}
\label{subfig:scheme}
 \end{figure}

\begin{equation}\label{eq:eq}
\begin{aligned}
g_1 = \hat{f}\left(x_n+\frac{h}{4}(g_1-g_2)\right), \\
g_2 = \hat{f}\left(x_n+ h\left(\frac{1}{4}g_1+\frac{5}{12}g_2\right)\right).
\end{aligned}
\end{equation}
After sampling $z$ and $w$, the RK scheme \eqref{eq:eq} is transformed into a minimization problem \eqref{eq:fullEQ}.
With $u_1=x_n+\frac{h}{4}(g_1-g_2)$, $u_2=x_n+h(\frac{1}{4}g_1+\frac{5}{12}g_2)$, and $\hat{z} = z-f_z$ it holds that $\begin{pmatrix}
g^{\star}_1 \\
g^{\star}_2
\end{pmatrix} = \arg \min_g F(g),$ with
\begin{equation}
\begin{aligned}
F(g)=\left\Vert\,\begin{matrix} g_1- \sum_{i=1}^{S} w_i \phi_i(u_1)-\sum_{j=1}^M v_j k(u_1,\xi_j)\hat{z} \\
g_2- \sum_{i=1}^{S} w_i \phi_i(u_2)-\sum_{j=1}^M v_j k(u_2,\xi_j)\hat{z}
\end{matrix}\,\right\Vert^2.
\end{aligned}
\end{equation}

\subsection{Application to model learning via variational inference}
\label{section:varInference}
Next, we construct a variational-inference based model learning scheme that is based on the previously introduced numerical integrators.
Here, we exemplary develop the integrators for an inference scheme similar to \citet{doerr2018probabilistic} and make the method precise.
It is also possible to extend the arguments to other inference schemes such as \citet{IalWilHenRas19, elef}.
In contrast to \citet{doerr2018probabilistic} we sample functions $\hat{f}$ instead of independent draws from the GP dynamics.
Thus, the produced trajectory samples refer to a probabilistic, but fixed vector field.
Unlike typical state-space models as \citet{doerr2018probabilistic, IalWilHenRas19} we omit transition noise.
Thus, the proposed variational posterior is suitable \citep{IalWilHenRas19}.
Structure preservation is in general not possible when adding transition noise to each time step \parencite[\textsection 5]{Abdulle_2020}.

Factorizing the joint distribution of noisy observations, noise-free states, inducing targets and GP posterior yields
% $p(\hat{x}_{1:N},x_{1:N},z,\hat{f})$
\begin{equation}\label{eq:ssm}
\begin{aligned}
&p(\hat{x}_{1:N},x_{1:N},z,\hat{f})\\
= \prod_{n=0}^{N-1} &p(\hat{x}_{n+1}|x_{n+1}) p(x_{n+1}|x_n,\hat{f})
p(\hat{f}|z) p(z).
\end{aligned}
\end{equation}
The posterior distribution $p(x_n, z, \hat{f}|\hat{x}_n)$ is factorized and approximated by a variational distribution $q(x_n,z,\hat{f})$.
Here, the variational distribution $q$ is chosen as
\begin{equation}\label{eq:var-dist}
\begin{aligned}
q(x_{1:N},z, \hat{f})&=\prod_{n=0}^{N-1} p(x_{n+1}|x_n,\hat{f}) p(\hat{f}|z) q(z),
\end{aligned}
\end{equation}
with the variational distribution $q(z)$ of the inducing targets from Section \ref{section:varGP}.
The model is adapted by maximizing the Evidence Lower Bound (ELBO)
\begin{equation}\label{eq:11}
\begin{aligned}
&\log p(\hat{x}_{1:N}) \geq \mathbb{E}_{q(x_{1:N},z, \hat{f})} \left[\log \frac{p(\hat{x}_{1:N},x_{1:N},z,\hat{f})}{{q(x_{1:N},z, \hat{f})}}\right] \\ =&\sum_{n=1}^{N} \mathbb{E}_{q(x_{1:N},z, \hat{f})}  \left[ \log p(\hat{x}_n|x_n) \right]
-\mathrm{KL}(p(z)||q(z))
=: \mathcal{L}.
\end{aligned}
\end{equation}

Now, the model can be trained by maximizing the ELBO $\mathcal{L}$ \eqref{eq:11} with a sampling-based stochastic gradient descent method that optimizes the sparse inputs and hyperparameters. The expectation $\mathbb{E}_{q(x_{1:N},z, \hat{f})} \left[\log \frac{p(\hat{x}_{1:N},x_{1:N},z,\hat{f})}{{q(x_{1:N},z, \hat{f})}}\right]$ is approximated by drawing samples from the variational distribution $q(\hat{x}_{1:N},z,\hat{f})$ and evaluating $p(\hat{x}_n|x_n)$ at these samples. 
Samples from $q$ are drawn by first sampling pseudo targets $z$ and a dynamics function $\hat{f}$ from the GP posterior \eqref{eq:matheron}.
Trajectories are produced by succesively computing consistent integrator steps $x_{n+1} = \psi_{\hat{f}}(x_n)$ as described in Sec. \ref{section:reparam}.
This yields a recurrent learning scheme, by iterating over multiple integration steps depicted in Figure~\ref{subfig:scheme}.
We are able to use our model for predictions by sampling functions from
the trained posterior.

\subsection{Gradients}
\label{subsection:grad}
The ELBO \eqref{eq:11} is minimized by applying stochastic gradient descent to the hyperparameters.
When conditioning on the sparse GP \eqref{eq:sparse-moments}, the hyperparameters include $\theta = (\mu_{1:d},\Sigma_{1:d},\theta_{1:d}^{GP})$ with variational sparse GP parameters $\mu_{1:d}$, $\Sigma_{1:d}$ and GP hyperparameters $\theta_{1:d}^{GP}$.
The gradient $\frac{d x_{n+1}}{d \theta}$ depends on $\frac{dk^{\star}}{d\theta}$ and $\frac{dx_n}{d\theta}$ via the integrator \eqref{eq:rk}. It holds that
\begin{equation} \label{eq:dkstar}
\frac{dg^{\star}}{d \theta} = \frac{\partial g^{\star}}{\partial \theta}+\frac{dg^{\star}}{d  x_n}\frac{d  x_n}{d\theta}.
\end{equation}
By the dependence of $x_{n+1}$ on $g^{\star}$ and of $g^{\star}$ on $x_n$, \eqref{eq:dkstar} the gradient is backpropagated through time. For an explicit integrator, the gradient $\frac{dk}{d\theta}$ can be computed explicitly, since $g^{\star}_j$ depends on $g^{\star}_i$ with $i < j$. For implicit solvers, the implicit functions theorem \citep{IFT} is applied.
It holds that $g^{\star} = \arg \min_{g} \eta(k, x_n, \theta)$ with the minimization problem $\eta$ derived in \eqref{eq:fullEQ}.
For the gradients of $g^{\star}$ with respect to $x_n$ respectively $\theta$ it holds with the implicit function theorem \citep{IFT}
\begin{equation}
\frac{dg^{\star}}{d  x_n}=\left(\frac{\partial ^2 \eta }{\partial {g^{\star}}^2}\right)^{-1}\left(\frac{\partial^2 \eta}{\partial  x_n \partial g^{\star}} \right).
\end{equation}

% !TEX root = aistats_2022.tex

\section{Application to symplectic integrators}
\label{section:app}
In summary, we have first derived how to evaluate the implicitly defined RK distributions.
Afterward, we have embedded this technique into a recurrent learning scheme and finally, shown how it is trained.
Next, we make the method precise for symplectic integrators and Hamiltonian systems.
\subsection{Hamiltonian systems and symplectic integrators}
\label{section:Hamiltonian}

An autonomous Hamiltonian system is given by
\begin{equation} \label{eq:hamiltonian}
x(t) = \begin{pmatrix}p(t)\\ q(t)\end{pmatrix} \textrm{ with } \dot{x}(t) = \begin{pmatrix}\dot{p}(t) \\ \dot{q}(t) \end{pmatrix}=\begin{pmatrix}-H_q(p,q) \\ H_p(p,q)  \end{pmatrix}
\end{equation}
and $p, q \in \mathbb{R}^d$.
In many applications, $q$ corresponds to the state and $p$ to the velocity.
The Hamiltonian $H$ often resembles the total energy and is constant along trajectories.
The flow of Hamiltonian systems $\psi_t$ is volume preserving in the sense of $\textrm{vol}(\psi_t(\Omega))=\textrm{vol}(\Omega)$ for each bounded open set $\Omega$.
The flow $\psi_t$ describes the solution at time point $t$ for the initial values $x_0 \in \Omega$.

Symplectic integrators are volume preserving for Hamiltonian systems \eqref{eq:hamiltonian} \citep{hairer2006geometric}. Thus, $\textrm{vol}(\Omega) = \textrm{vol}(\psi_{f}(\Omega))$ for each bounded $\Omega$.
Further, they provide a more accurate approximation of the total energy than standard integrators \citep{hairer2006geometric}.
When designing the GP, it is critical to respect the Hamiltonian structure \eqref{eq:hamiltonian}.
Additionally, the symplectic integrator ensures that the volume is indeed preserved.
\subsection{Explicit symplectic integrators}
\label{ssec:sep}
A broad class of real world systems can be modeled by separable Hamiltonians $H(p,q)=T(p)+V(q)$.
For example, ideal pendulums and the two body problem.
Then, for the dynamical system it holds that
\begin{equation}  \label{eq:sep-ham}
\dot{p}(t) =-V^\prime (q), \,\dot{q}(t)=T^\prime (p),
\end{equation}
with $V: \mathbb{R}^d \rightarrow \mathbb{R}^d$ and $T: \mathbb{R}^d \rightarrow \mathbb{R}^d$.
For this class of problems, explicit symplectic integrators can be constructed.
In order to ensure Hamiltonian structure, $V ^{\prime}_1(q), \dots, V ^{\prime}_d(q)$ and $ T ^{\prime}_1(p), \dots, T ^{\prime}_d(p)$ are modeled with independent sparse GPs.
Symplecticity is enforced via discretizing with a symplectic integrator.

Consider for example the explicit symplectic Euler method
\begin{equation} \label{eq:symp-euler}
p_{n+1}=p_n-hV^{\prime}(q_n), \, q_{n+1}=q_n+hT^{\prime}(p_{n+1}).
\end{equation}
The symplectic Euler method \eqref{eq:symp-euler} is a partitioned RK method, meaning that different schemes are applied to different dimensions.
Here, the explicit Euler method is applied to $p_n$ and the implicit Euler method to $q_n$.
The method \eqref{eq:symp-euler} is reparametrized as described in Sec.~\ref{section:reparam} by sampling from $V^{\prime}$ and $T^{\prime}$ and the scheme can readily be embedded into the inference scheme (\cf Sec. \ref{section:varInference}).

\subsection{General symplectic integrators}
\label{ssec:nonsep}
The general Hamiltonian system \eqref{eq:hamiltonian} requires the application of an implicit symplectic integrator.
An example for a symplectic integrator is the midpoint rule applied to \eqref{eq:hamiltonian}
\begin{equation} \label{eq:midpoint}
\begin{aligned}
x_{n+1}&=x_n+h J^{-1} \nabla H \left(\frac{x_n+x_{n+1}}{2}\right), \\
\textrm{ with }
J^{-1}&= \left( \begin{matrix}
0 & -1 \\
1 & 0
\end{matrix} \right).
\end{aligned}
\end{equation}
Again, it is critical to embed the Hamiltonian structure into the dynamics model by modeling $H$ with a sparse GP.
Sampling from \eqref{eq:midpoint} requires evaluating the gradient $\nabla H$, which is again a GP \citep{10.5555/1162254, inbook}.

% !TEX root=aistats_2022.tex
\section{Experiments}
\label{sec:exp}
In this section, we validate our method numerically. In particular, we show that we i) achieve higher accuracy than comparable state-of-the-art methods;
ii) demonstrate volume-preserving predictions for Hamiltonian systems and the satisfaction of a quadratic geometric constraint; and iii) illustrate that our method can easily deal with different choices of RK integrators.

\subsection{Methods}
We construct our structure-preserving GP model (SGPD) by tailoring the RK integrator to the underlying problem. We compare to the following state-of-the-art approaches:

\textbf{Hamiltonian neural network (HNN)} \parencite{NIPS20199672}: Deep learning approach that is tailored to respect Hamiltonian structure.

\textbf{Consistent PR-SSM model (Euler)} \cite{doerr2018probabilistic}: Standard variational GP model that corresponds to explicit Euler discretizations. Therefore, we refer to it in the following as Euler. In general all common GP state-space models correspond to the Euler discretization. Here, we use a model similar to \cite{doerr2018probabilistic}, but in contrast to \cite{doerr2018probabilistic} we compute consistent predictions via decoupled sampling in order to provide the nescessary comparability and since the inconsistent sampling scheme in \cite{doerr2018probabilistic} was critized in later work \citep{IalWilHenRas19}.
The general framework is more flexible and can also cope with lower-dimensional state observations. Here, we consider the special case, where we assume noisy state measurements.
\subsection{Learning task and comparison}
In the following, we describe the common setup of all experiments. For each Hamiltonian system, we consider one period of a trajectory as training data and all methods are provided with identical training data.
For all experiments, we choose the ARD kernel \citep{10.5555/1162254}. We apply the training procedure described in Sec. \ref{section:varInference} on subtrajectories and perform predictions via sampling of trajectories.
In order to draw a fair comparison, we choose similar hyperparameters and number of inducing inputs for our SGPD method and the standard Euler discretization. Details are moved to the appendix.
In contrast to our method the HNN requires additional derivative information, either analytical or as finite differences.
Here, we assume that analytical derivative information is not available and thus compute finite differences.

We consider a twofold goal: accurate predictions in the $L^2$ sense and invariant preservation.
Predictions are performed by unrolling the first training point over multiple periods of the system trajectory. The $L^2$-error is computed via averaging 5 independent samples $\hat{X}_i = \frac{1}{5}\sum_{j=1}^5 \hat{X}_i^j$ and computing $\sqrt{\frac{1}{N} \sum_{i=1}^N {\Vert X_i-\hat{X}_i \Vert^2}}$ with ground truth $X$.
Integrators are volume-preserving if and only if they are divergence free, which requires  $\det(\psi')=1$ \citep{hairer2006geometric}.
Thus, we evaluate $\det(\psi')$ for the rollouts, which is intractable for the Hamiltonian neural networks.
We observed that we can achieve similar results by propagating the GP mean in terms of constraint satisfaction and $L^2$-error.

\begin{figure*}[h!]
	% \centering
	% \hfill
	\begin{subfigure}[htb]{0.5\textwidth}
		\centering
		\includegraphics[width=0.7\textwidth]{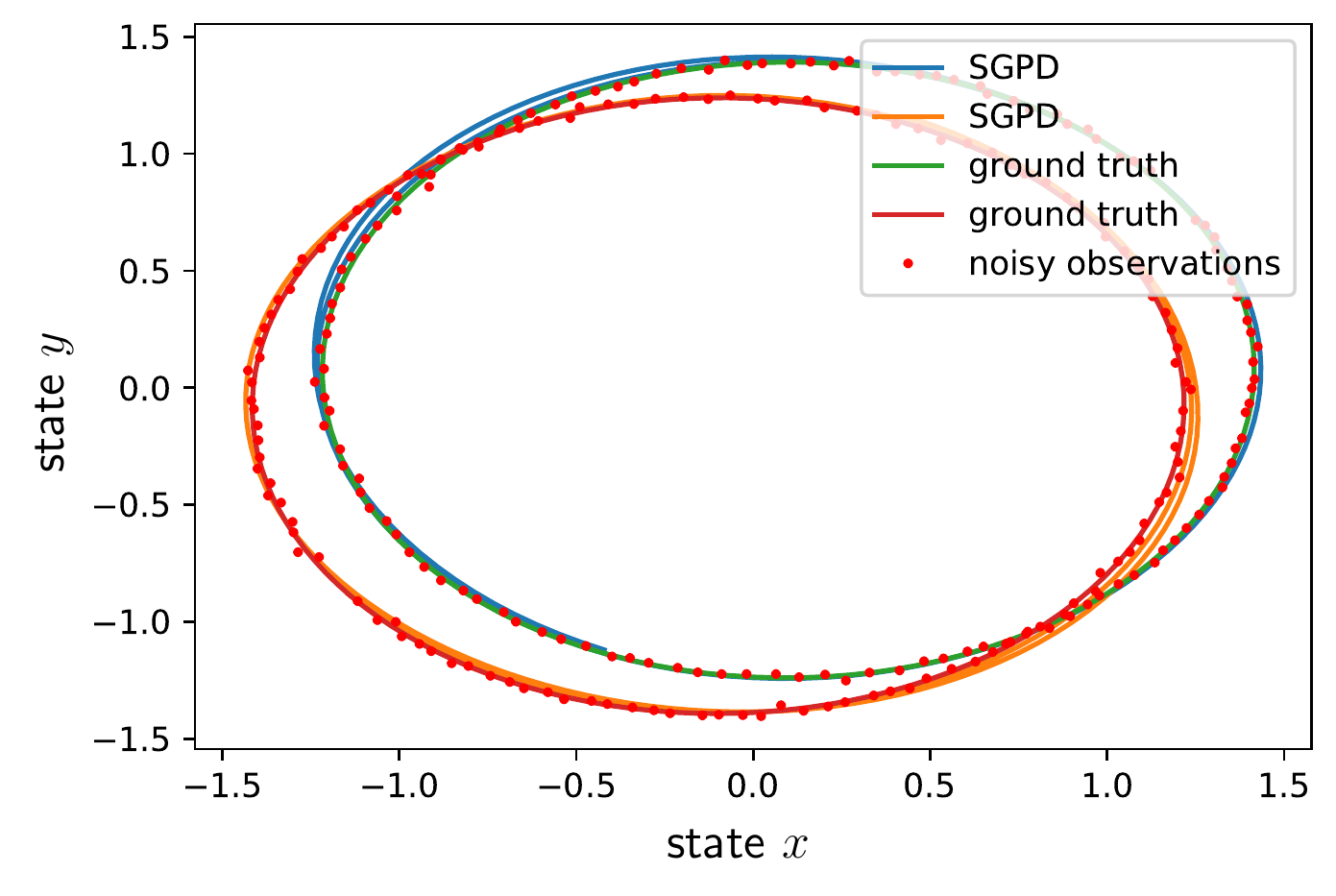}
	\end{subfigure}
	% \hfill
	\begin{subfigure}[htb]{0.5\textwidth}
		\centering
		\includegraphics[width=0.7\textwidth]{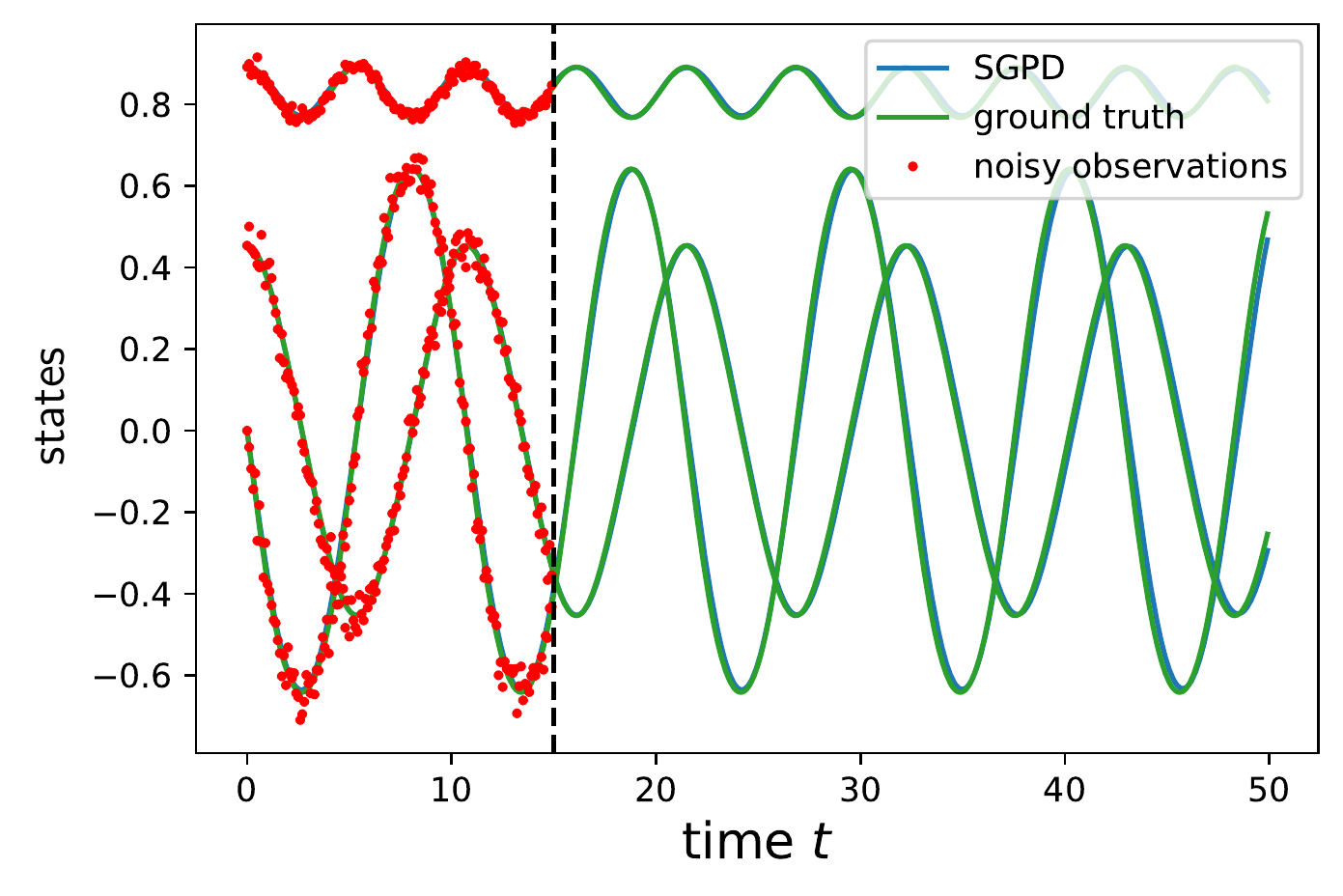}
		%	\caption{Rigid body motion} \label{subfig:RB}
	\end{subfigure}
	% \hfill
	\caption{State trajectories for the two-body problem (left) and rigid body dynamics (right). The rigid body dynamics are illustrated as a function over time to show the location of the training data and future behavior of the system. The two-body problem is represented as a phase plot in the two-dimensional space. The first particle $q_1$ is shown in green and $q_2$ in red.}
	\label{fig:sep}
\end{figure*}

\begin{table*}[ht]
	\caption{Shown are the total $L^2$-errors in \ref{t:errors} and an analysis of the total energy for the non-separable system \ref{t:nonsep}.}
	\begin{subtable}[c]{0.49\textwidth}
		% \flushleft
		\caption{total $L^2$-errors (mean (std) over 5 indep.\ runs)}
		\label{t:errors}
		\begin{tabular}{rccc}
			\noalign{\smallskip} \hline \hline \noalign{\smallskip}
			task & SGPD & Euler & HNN \\
			\hline
			(i) & \textbf{0.421} (0.1) & 0.459 (0.12) & 4.69 (0.02)\\
			(ii) & \textbf{0.056} (0.01)  & 0.057 (0.009)   & 0.12 (0.009) \\
			(iii) & \textbf{0.033} (0.01) & 0.034 (0.021) & 0.035 (0.007)\\
			(iv)  & \textbf{0.046} (0.014) & 0.073 (0.02) & - \\
			\noalign{\smallskip} \hline \noalign{\smallskip}
		\end{tabular}\quad
	\end{subtable}\;
	\begin{subtable}[c]{0.49\textwidth}
		\caption{Energy for non-separable Hamiltonian}
		% \flushright
		\centering
		\label{t:nonsep}
		\begin{tabular}{rcc}
			\noalign{\smallskip}  \hline \hline \noalign{\smallskip}
			method & energy err. & std. dev. \\
			\hline
			SGPD & $\mathbf{9\cdot 10^{-4}}$   & $\cdot 10^{-3}$ \\
			Euler & $10^{-3}$  & $4 \cdot 10^{-3}$  \\
			HNN & $10^{-3}$ & $2 \cdot 10^{-3}$ \\
			\noalign{\smallskip} \hline \noalign{\smallskip}
		\end{tabular}
	\end{subtable}
\end{table*}

\begin{figure*} 
	\begin{subfigure}[htb]{0.33\textwidth}
		\includegraphics[width=0.9\textwidth]{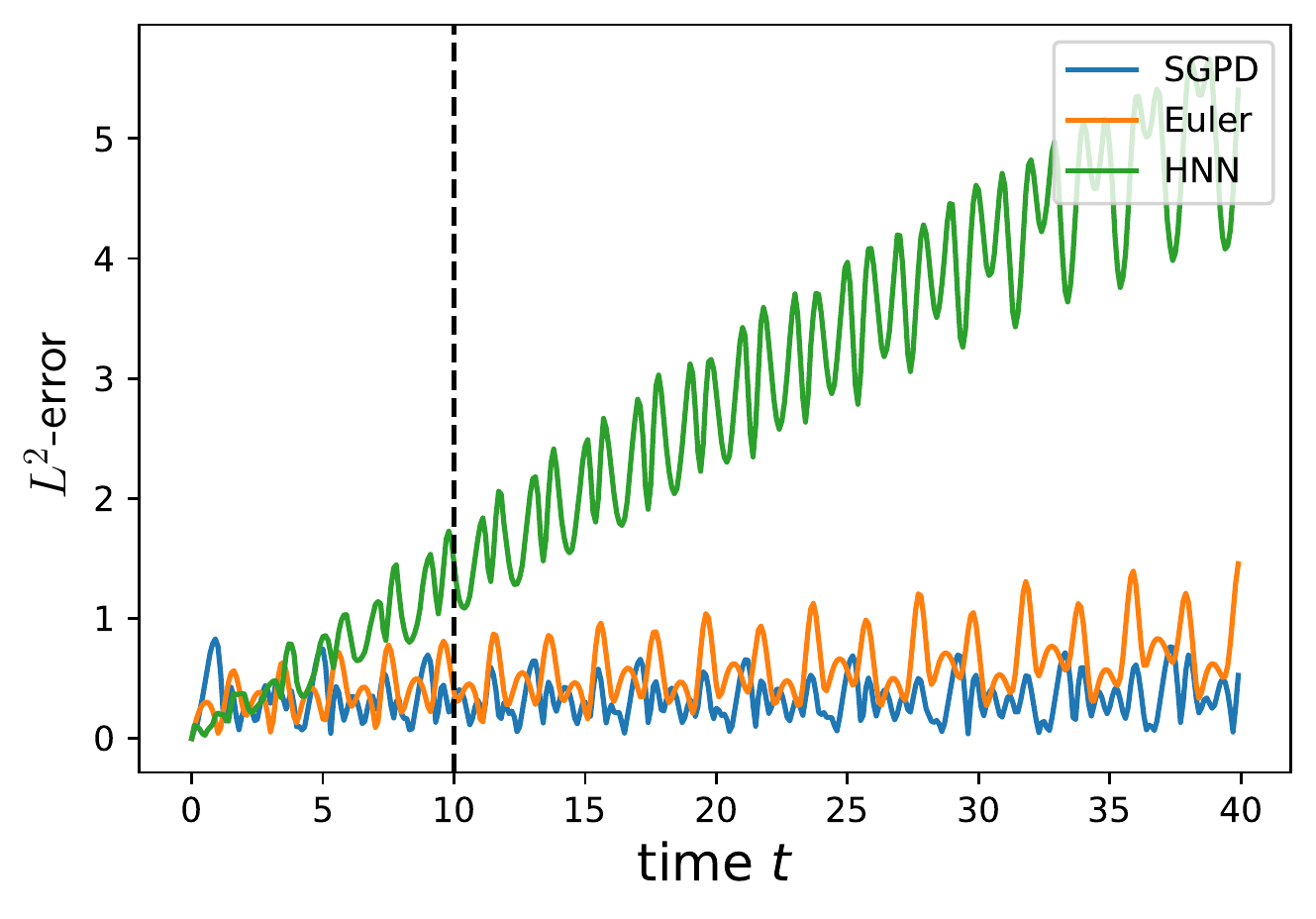}
	\end{subfigure}
	\begin{subfigure}[htb]{0.33\textwidth}
		\includegraphics[width=0.9\textwidth]{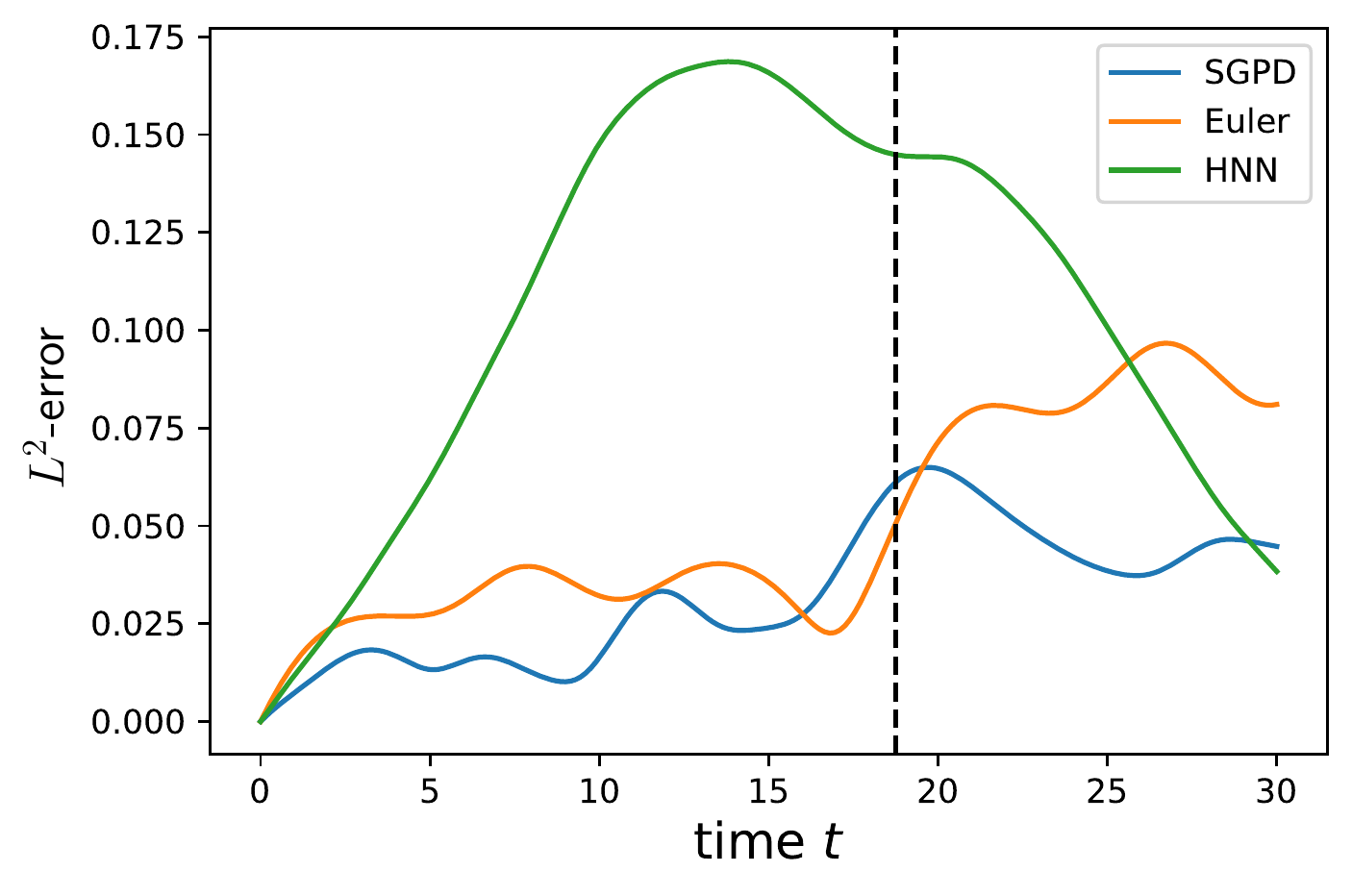}
	\end{subfigure}
	\begin{subfigure}[htb]{0.33\textwidth}
		\includegraphics[width=0.9\textwidth]{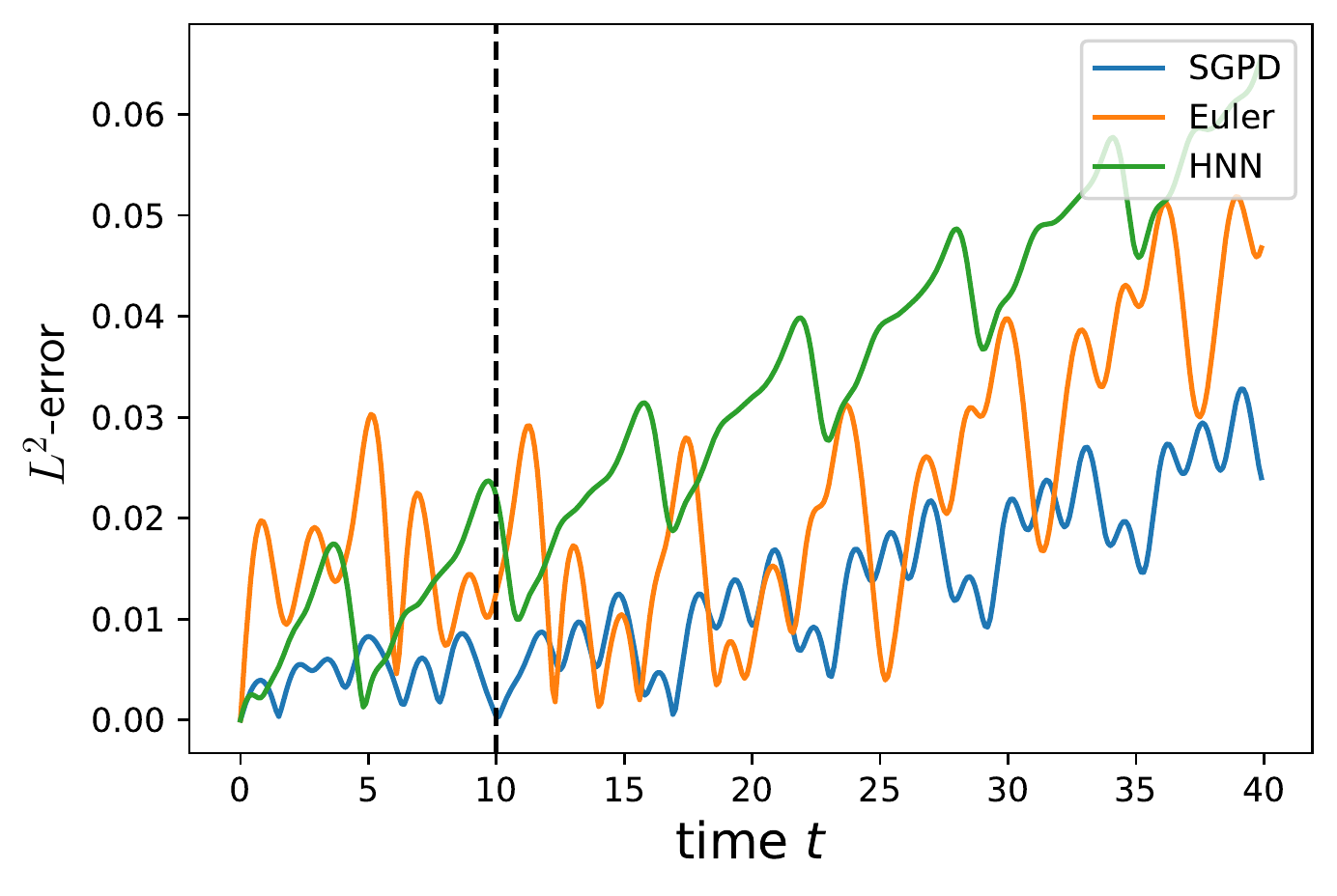}
	\end{subfigure}
	\caption{$L^2$-errors of averaged state trajectories for all three methods.
		Shown are the errors for the pendulum (left), the two-body problem (middle), and the non-separable Hamiltonian (right).
		The training horizon is marked.}
	\label{fig:l2}
\end{figure*}

\begin{figure*}[h!]
	\begin{subfigure}[t]{0.33\textwidth}
		\includegraphics[width=\textwidth]{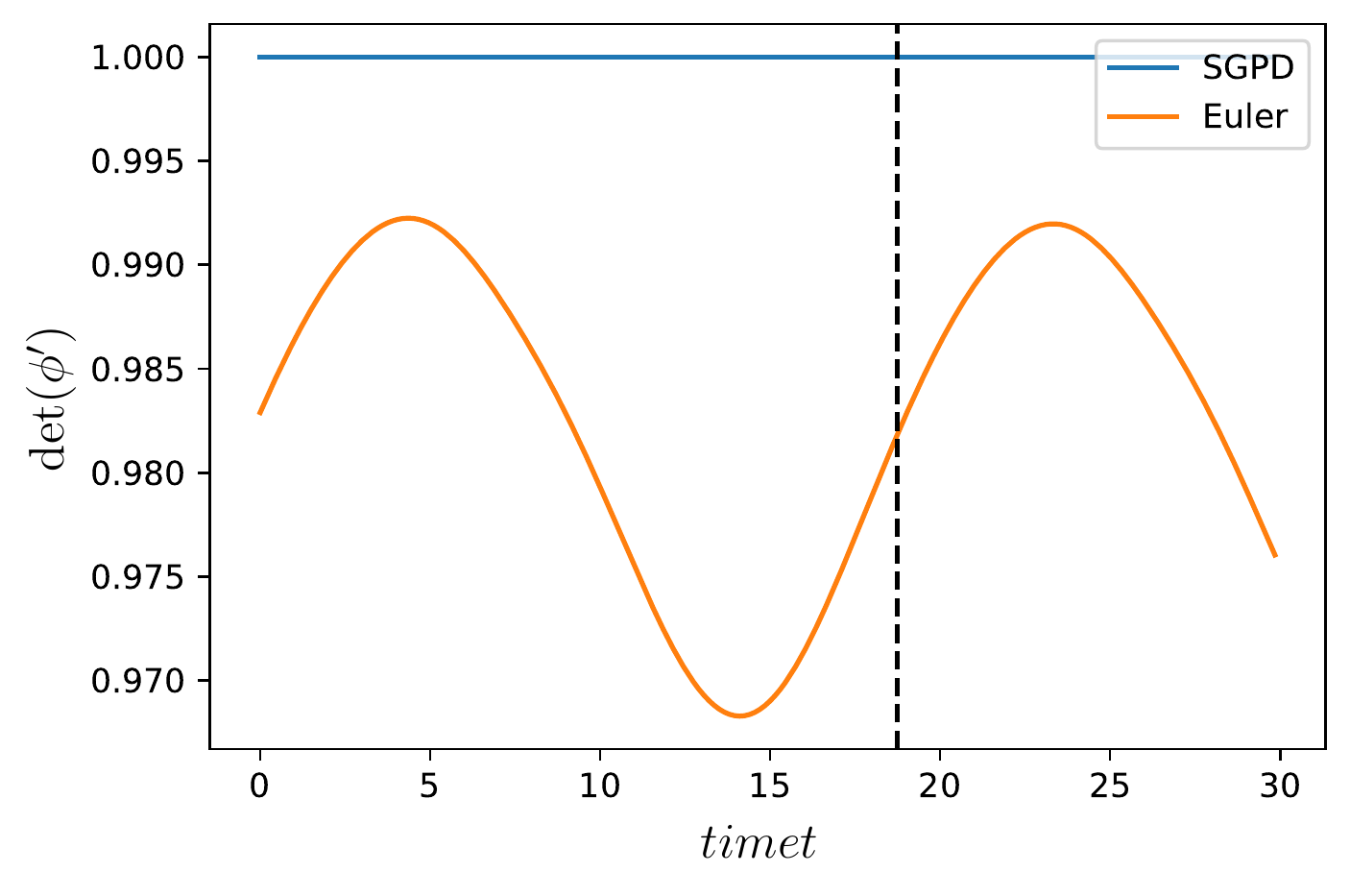}
		\caption{Two-body problem}\label{subfig:TwoBodyDet}
	\end{subfigure}
	\begin{subfigure}[t]{0.33\textwidth}
		\includegraphics[width=\textwidth]{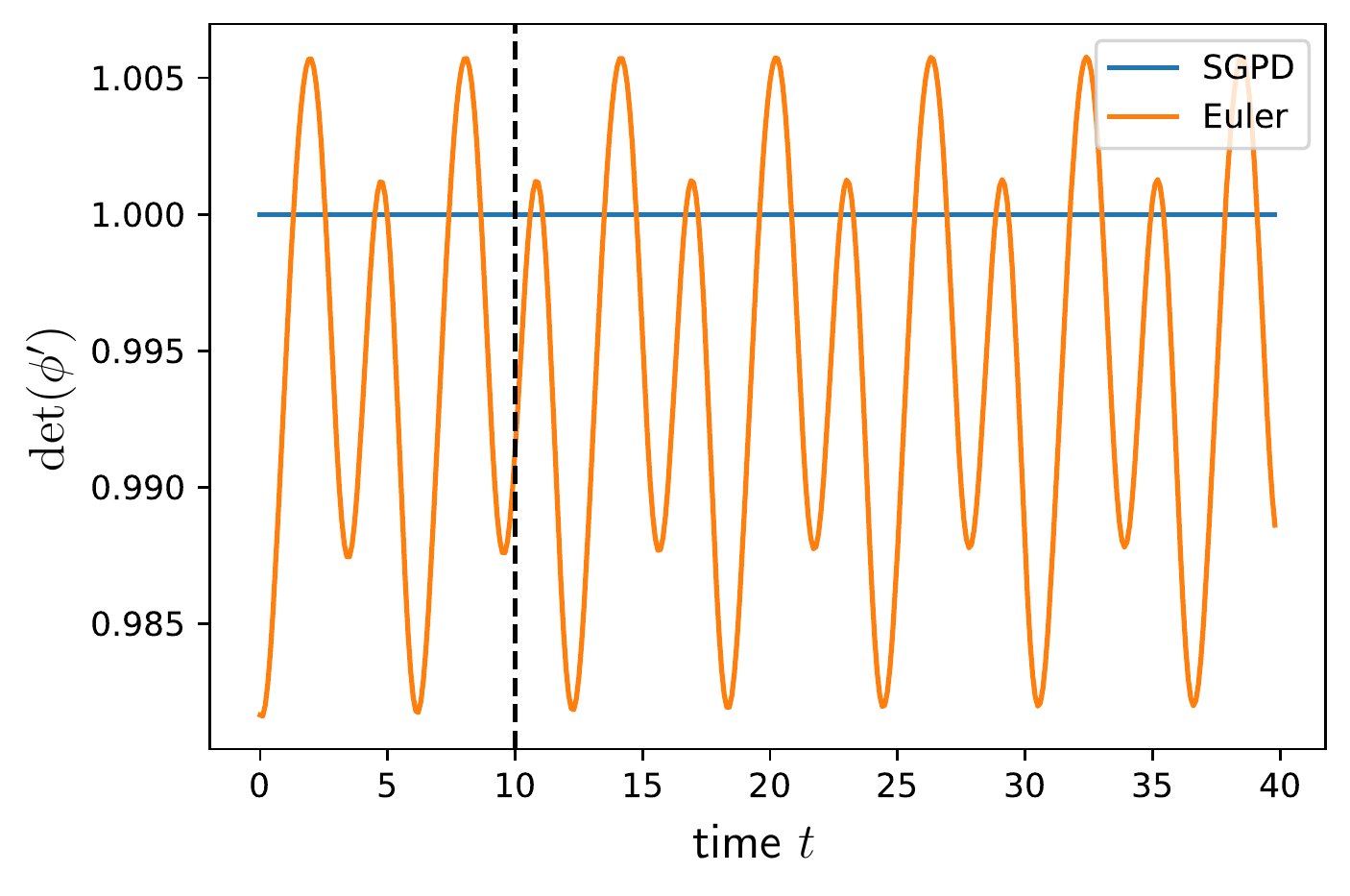}
		\caption{Non-separable system} \label{subfig:implicitDet}
	\end{subfigure}
	\begin{subfigure}[t]{0.33\textwidth}
		\includegraphics[width=\textwidth]{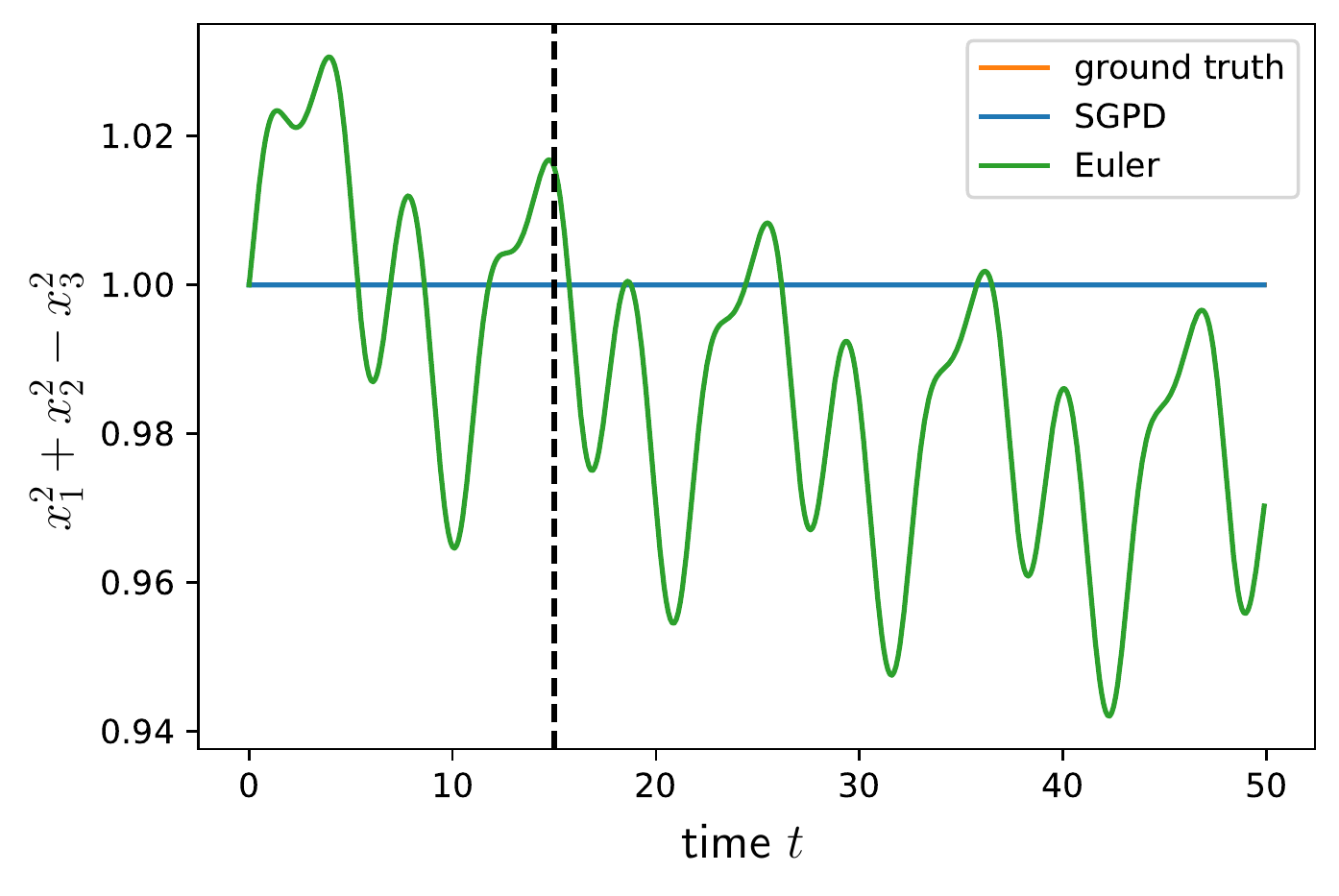}
		\caption{Rigid body dynamics}\label{subfig:RBEnergy}
	\end{subfigure}
	\caption{The proposed SGPD (blue) preserves structure: The analytic volume is preserved via SGPD for the Hamiltonian systems in \ref{subfig:TwoBodyDet} and \ref{subfig:implicitDet}, while simulations show that the explicit Euler (orange) does not preserve volume. Further, the SGPD rollout preserves the quadratic constraint over time for the rigid body system in \ref{subfig:RBEnergy}, while the standard GP with explicit Euler does not.
	}	\label{fig:invariants}
\end{figure*}

\subsection{Systems and integrators}
We consider four different systems here i) ideal pendulum; ii) two-body problem; iii) non-separable Hamiltonian; and iv) rigid body dynamics.

\textbf{Separable Hamiltonians:} the systems i) and ii) are both separable Hamiltonians that are also considered as baseline problems in \citet{NIPS20199672}. Due to the similar structure, we can apply the symplectic Euler method (\cf Sec. \ref{ssec:sep}) to both problems.

The Hamiltonian of a pendulum is given by $H(p,q)=(1-6\cos(p))+\frac{p^2}{2}$.
Training data is generated from a 10 second ground truth trajectory with discretization $dt=0.1$ and disturbed with observation noise with variance $\sigma^2=0.1$.
Predictions are performed on a 40 second interval.

The two body problem models the interaction of two unit-mass particles $(p_1, q_1)$ and $(p_2, q_2)$, where $p_1, p_2, q_1, q_2 \in \mathbb{R}^2$ and $H(p,q)=\frac{1}{2}+\Vert p_1 \Vert ^2 +\Vert p_2 \Vert^2+\frac{1}{\Vert q_1-q_2\Vert^2}$.
Noisy training data is generated on an interval of 18.75 seconds, discretization level $dt = 0.15$, and variance $\sigma^2= 1\cdot 10^{-3}$.
Predictions are performed on an interval of 30 seconds.
The orbits of the two bodies $q_1$ and $q_2$ are shown in Figure \ref{fig:sep} (left).

\textbf{Non-separable Hamiltonian:}
As an example for a non-separable Hamiltonian system we consider Eq. \eqref{eq:hamiltonian} with $H(p,q)=\frac{1}{2} \left[(q^2+1)(p^2+1) \right]$ \citep{Tao_2016}.
The implicit midpoint rule \eqref{eq:midpoint} is applied as the numerical integrator (\cf Sec. \ref{ssec:nonsep}).
The training trajectory is generated on a 10 seconds interval with discretization $dt = 0.1$ and disturbed with noise with variance $\sigma^2=5 \cdot 10^{-4}$.
Rollouts are performed on an interval of 40 seconds.

\textbf{Rigid body dynamics:}
Consider the rigid body dynamics  \citep{hairer2006geometric}
\begin{equation} \label{eq:RB}
\begin{pmatrix}
\dot{x}_1 \\
\dot{x}_2\\
\dot{x}_3
\end{pmatrix}
=\begin{pmatrix}
0 & \frac{3}{2}x_3 & -x_2 \\
-\frac{3}{2}x_3 & 0 & \frac{x_1}{2} \\
x_2 & - \frac{x_1}{2} & 0
\end{pmatrix}
\begin{pmatrix}
x_1 \\
x_2\\
x_3
\end{pmatrix} =:f(x)
\end{equation}
that describe the angular momentum of a body rotating around an axis.
The equations of motion can be derived via a constrained Hamiltonian system.
We apply the implicit midpoint method.
Since the HNN is designed for non-constrained Hamiltonians it requires pairs of $p$ and $q$ and is, thus, not applicable.
Training data is generated on a 15 seconds interval with discretization $dt=0.1$.
Due to different scales, $x_1$ and $x_2$ are disturbed with noise with variance $\sigma^2 = 1\cdot 10^{-3}$, and $x_3$ is disturbed with noise with variance $\sigma^2 = 1\cdot10^{-4}$.
Predictions are performed on an interval of 50 seconds (see Figure \ref{fig:sep} (right)). The rigid body dynamics preserve the invariant $x_1^2+x_2^2+x_3^2 = 1$, which refers to the ellipsoid determined by the axis of the rotating body.
We include this property as prior knowledge in our SGPD model via  $x^T \hat{f}(x)=1$ \citep{hairer2006geometric}.
The dynamics $\hat{f}$ is again trained with independent sparse GPs, where the third dimension is obtained by solving $\hat{f}(x)=1-\frac{\hat{f}_1 x_1 +\hat{f}_2x_2}{x_3}$.
% Appendix.

\subsection{Results}
In summary, our method shows the smallest $L^2$-error (see Table \ref{t:errors}).
The evolution of $L^2$-errors is illustrated in Figure \ref{fig:l2} for the pendulum (left), two-body problem (middle), and non-separable system (right) and show how the other methods accumulate errors.

For systems (i),(ii), and (iii), we demonstrate volume preservation.
Figure  \ref{fig:invariants} shows that volume is preserved for the symplectic integrator-based SGPD in contrast to the standard explicit Euler method.
For the rigid body dynamics, we consider the invariant $x_1^2+x_2^2+x_3^2 =1$.
Figure \ref{subfig:RBEnergy} demonstrates that the implicit midpoint is able to approximately preserve the invariant along the whole rollout.
In contrast, the explicit Euler fails even though it provides comparable accuracy in terms of $L^2$-error.

The midpoint method-based SGPD furthermore shows accurate approximation of the constant total energy for the systems (iii) and (iv).
The total energy corresponds to the Hamiltonian $H$ for system (iii).
We average the approximated energy along 5 independent trajectories $H_n = \sum_{i=1}^5 \frac{H_n^i}{5}$ and compute the average total energy $\hat{H}=\frac{1}{n}\sum_n H_n$. Afterwards we evaluate the error $\|H-\hat{H} \|$ and the standard deviation $\sqrt{\sum_n \frac{|H_n-H|^2}{n-1}}$ (see Table \ref{t:nonsep}).
Our SGPD method yields the best approximation to the energy.
Both methods outperform the explicit Euler method.
For the rigid body dynamics, our method yields extremly accurate approximation of the total energy. Details are moved to the Appendix.
An empirirical evaluation of higher-order methods is moved to the Appendix.

% !TEX root = aistats_2022.tex

\section{Conclusion and future work}
\label{sec:conclusion}
In this paper we combine numerical integrators with GP regression.
Thus, resulting in an inference scheme that preserves physical properties and yields high accuracy.
On a technical level, we derive a method that samples from implicitly defined distributions.
By the means of empiricial comparison, we show the advantages over Euler-based state-of-the-art methods that are not able to preserve physical structure.
Of course, our method critically relies on physical prior knowledge.
However, this insight is often available.
An important extension that we want to address in the future are state observations, control input, and continuous-time dynamics.

\clearpage
\bibliography{bib}

\end{document}

\onecolumn
\aistatstitle{Structure-preserving Gaussian Process Dynamics:\\
Supplementary Materials}

\section{Additional experiments} \label{sec:addExp}
We provide additional plots in order to illustrate the results from Section 6.
Furthermore we add uncertainty information to the SGPD predictions for all experiments in Section 6. 
We extend the results for the rigid body dynamics experiment by demonstrating the evolution of energy. 
In order to illustrate properties of higher-order integrators we provide additional experiments.
 
\begin{figure*}[h!]
	\begin{subfigure}[t]{0.33\textwidth}
		\includegraphics[width=\textwidth]{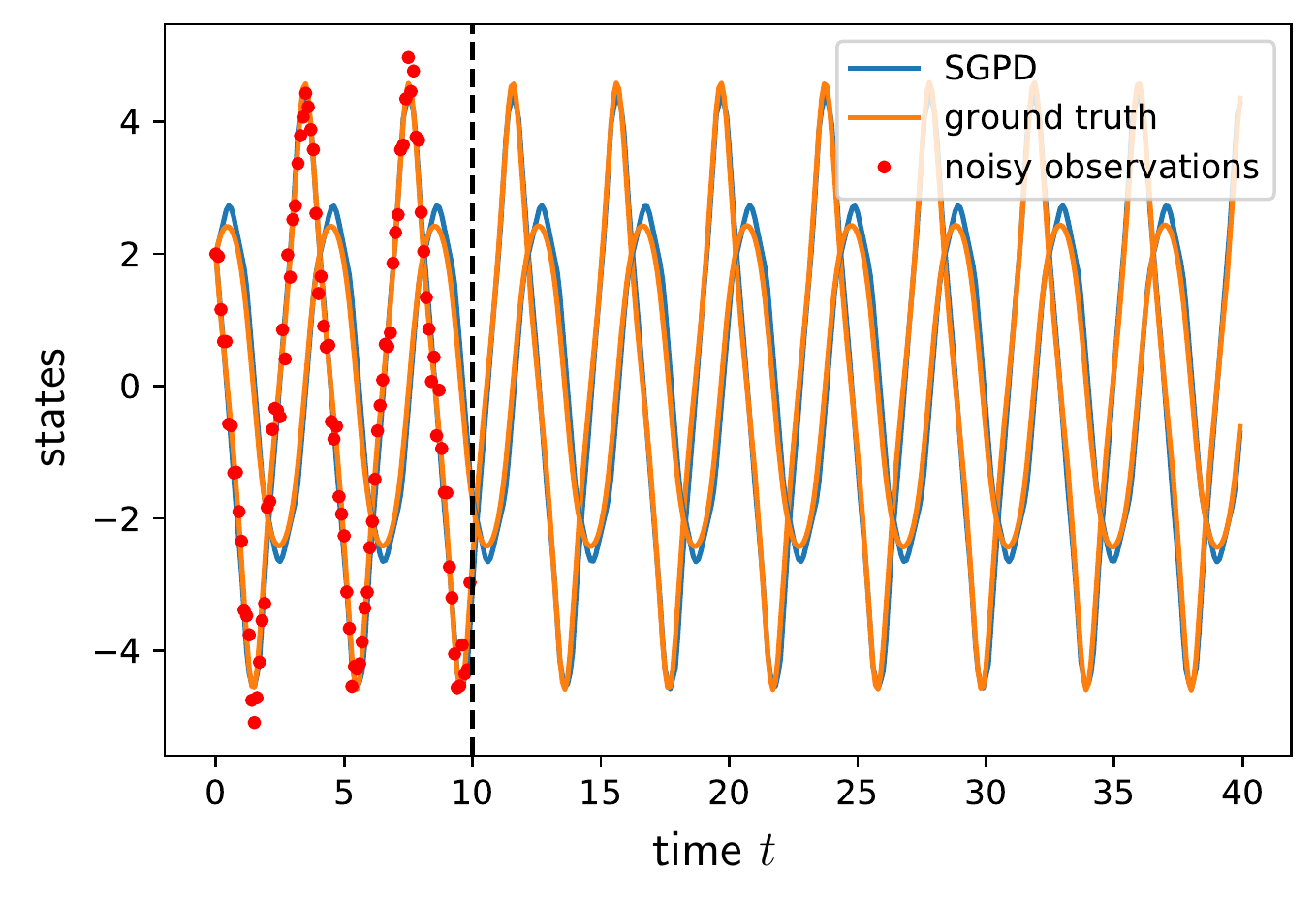}
	     \caption{SGPD rollout pendulum}\label{subfig:TwoBody}
	\end{subfigure}
	\begin{subfigure}[t]{0.33\textwidth}
		\includegraphics[width=\textwidth]{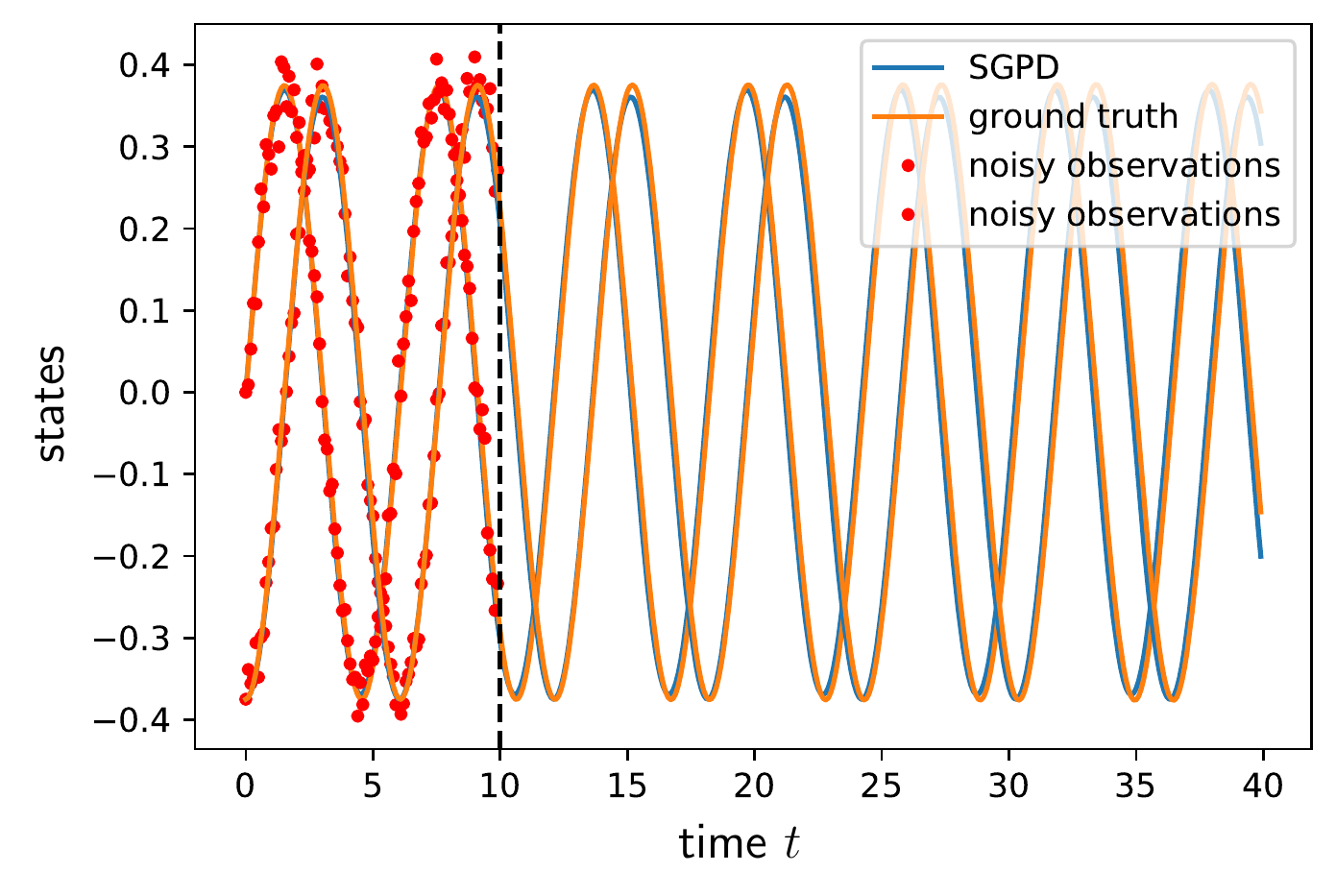}
		\caption{SGPD non-separable system} \label{subfig:RB}
	\end{subfigure}
	\begin{subfigure}[t]{0.33\textwidth}
		\includegraphics[width=\textwidth]{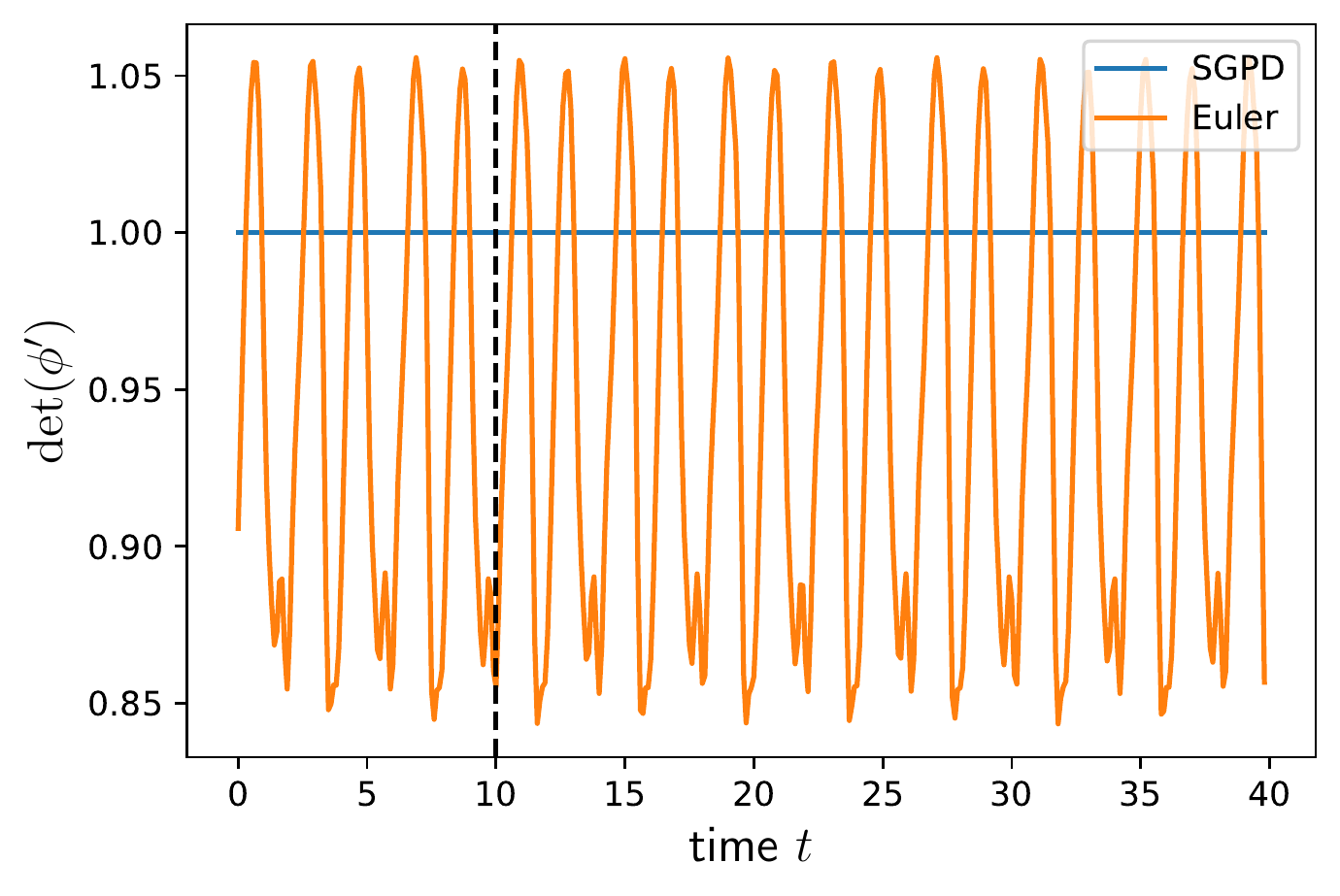}
		\caption{Pendulum determinant}\label{subfig:pendDet}
	\end{subfigure}
	\caption{Additional plots for the experiments in Section 6. Shown are state trajectories for the pendulum (left) and non-separable Hamiltonian system (middle). Both systems are illustrated as a function over time to show the location of the training data and future behavior of the system. The determinant of the pendulum problem is depicted (right) indicating that the explicit Euler method is not preserving structure.}		
	\label{fig:states}
\end{figure*}

Figure \ref{fig:states} shows additional plots for the results in Section 6. Shown are SGPD rollouts for the pendulum problem (left), the non-separable Hamiltonian system (middle), and the determinant of the pendulum problem (right).  

\subsection{Uncertainty} \label{ssec:uncertainty}
\begin{figure*}[tb]
	\begin{subfigure}[t]{0.49\textwidth}
		\includegraphics[width=\textwidth]{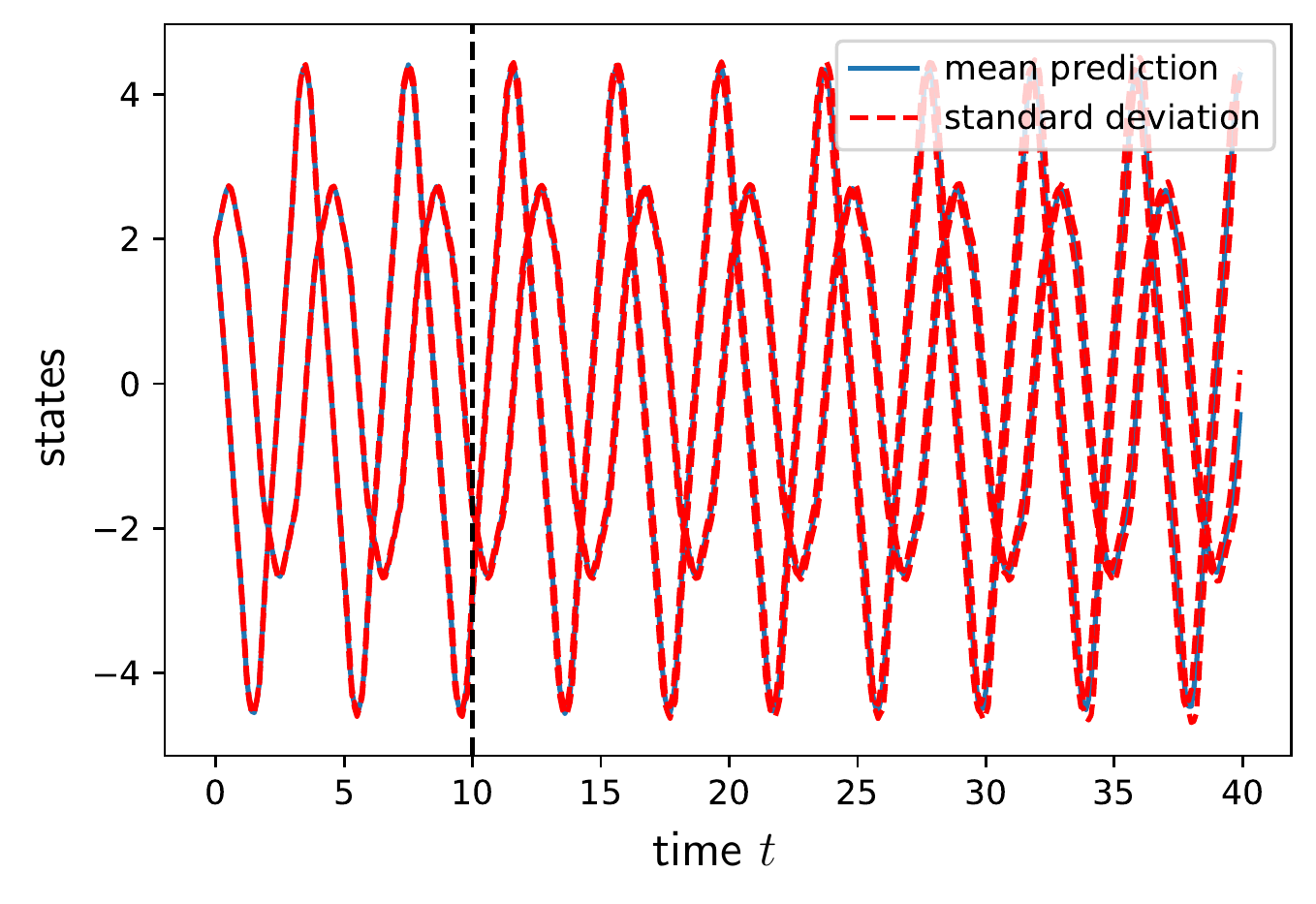}
		\caption{Pendulum uncertainty}\label{subfig:pendUncertainty}
	\end{subfigure}
	\begin{subfigure}[t]{0.49\textwidth}
		\includegraphics[width=\textwidth]{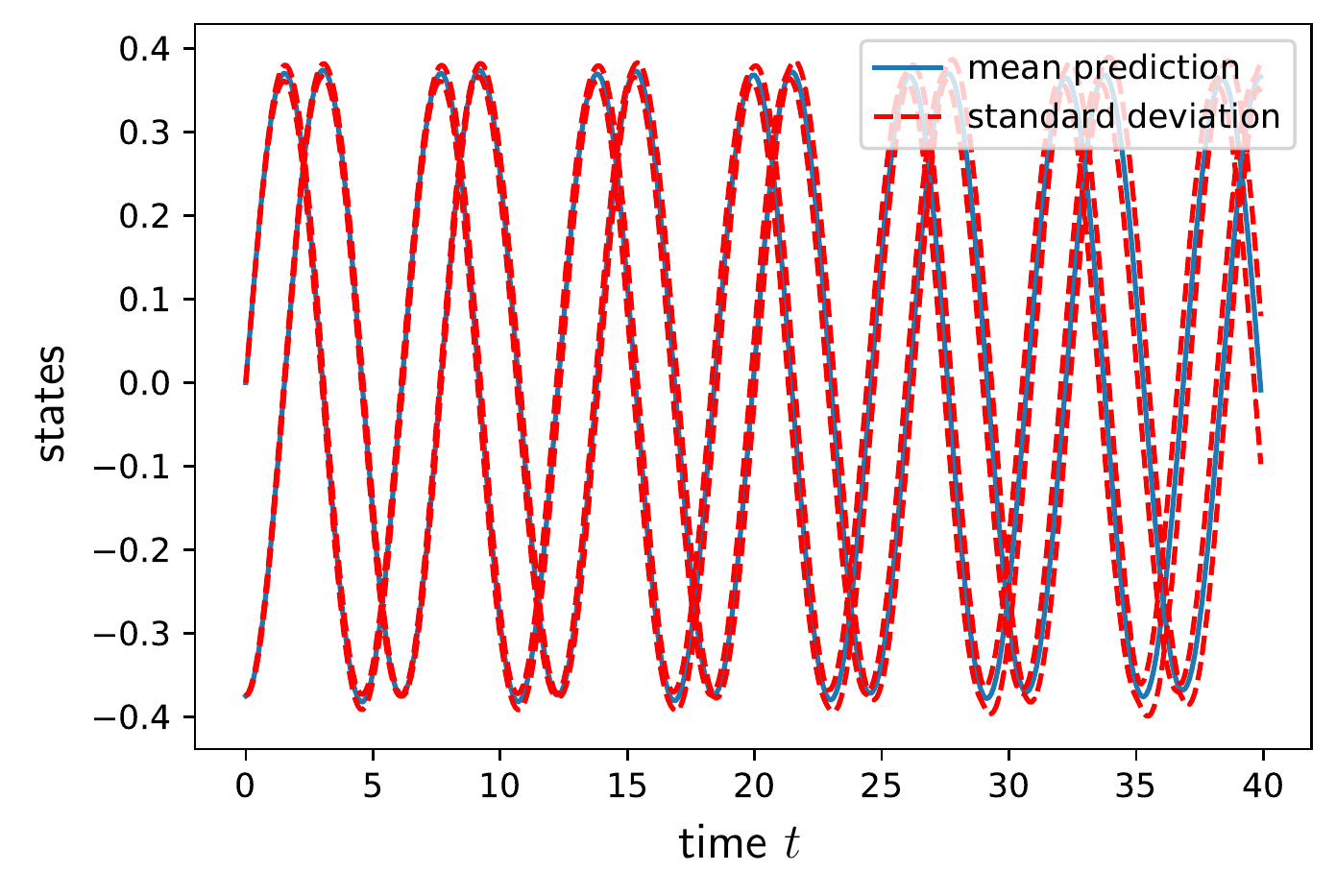}
		\caption{Non separable Hamiltonian uncertainty} \label{subfig:ToyProblemUncertainty}
	\end{subfigure}
	
	\begin{subfigure}[t]{0.49\textwidth}
		\includegraphics[width=\textwidth]{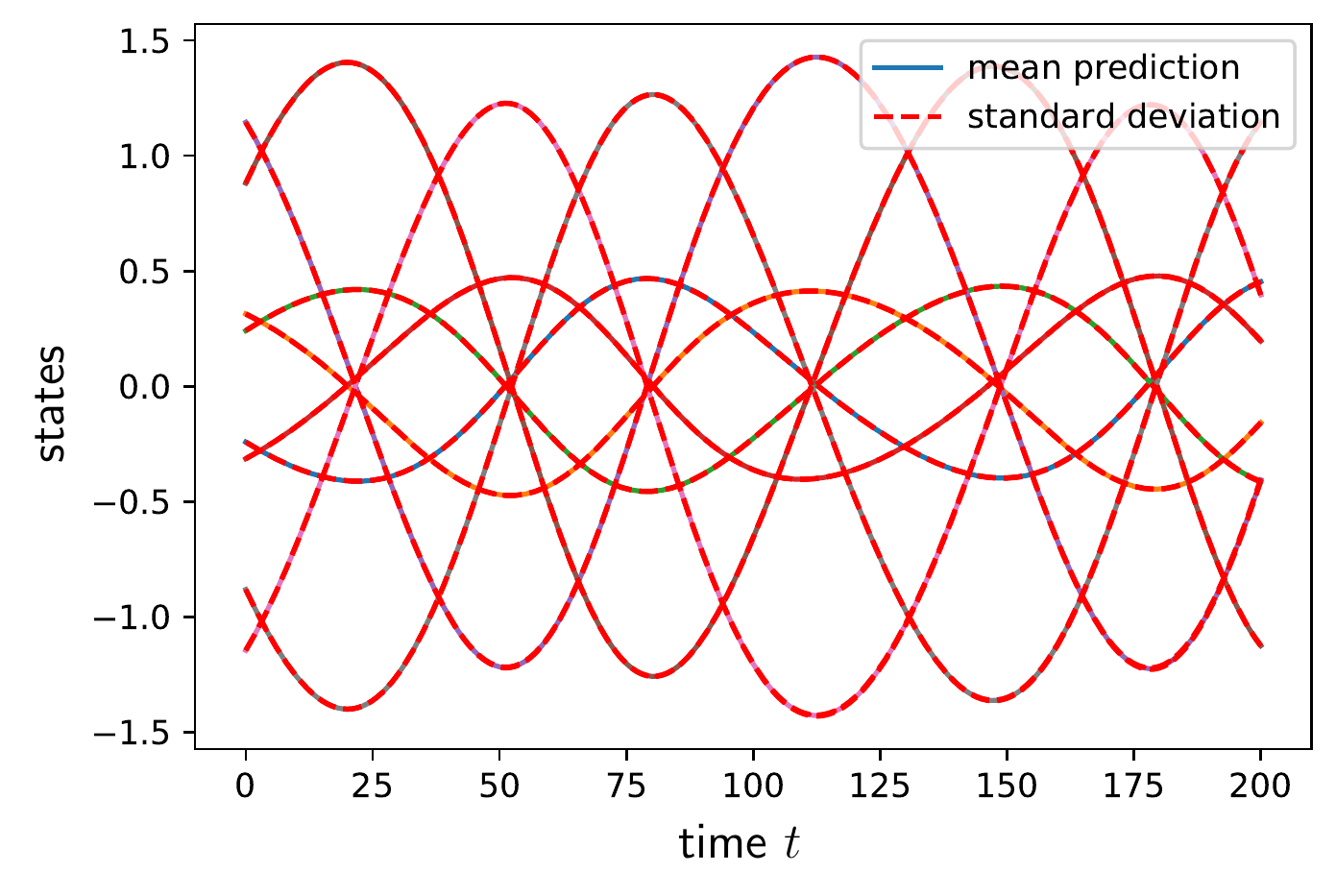}
		\caption{Two Body problem uncertainty}\label{subfig:twoBodyUncertainty}
	\end{subfigure}
	\begin{subfigure}[t]{0.49\textwidth}
		\includegraphics[width=\textwidth]{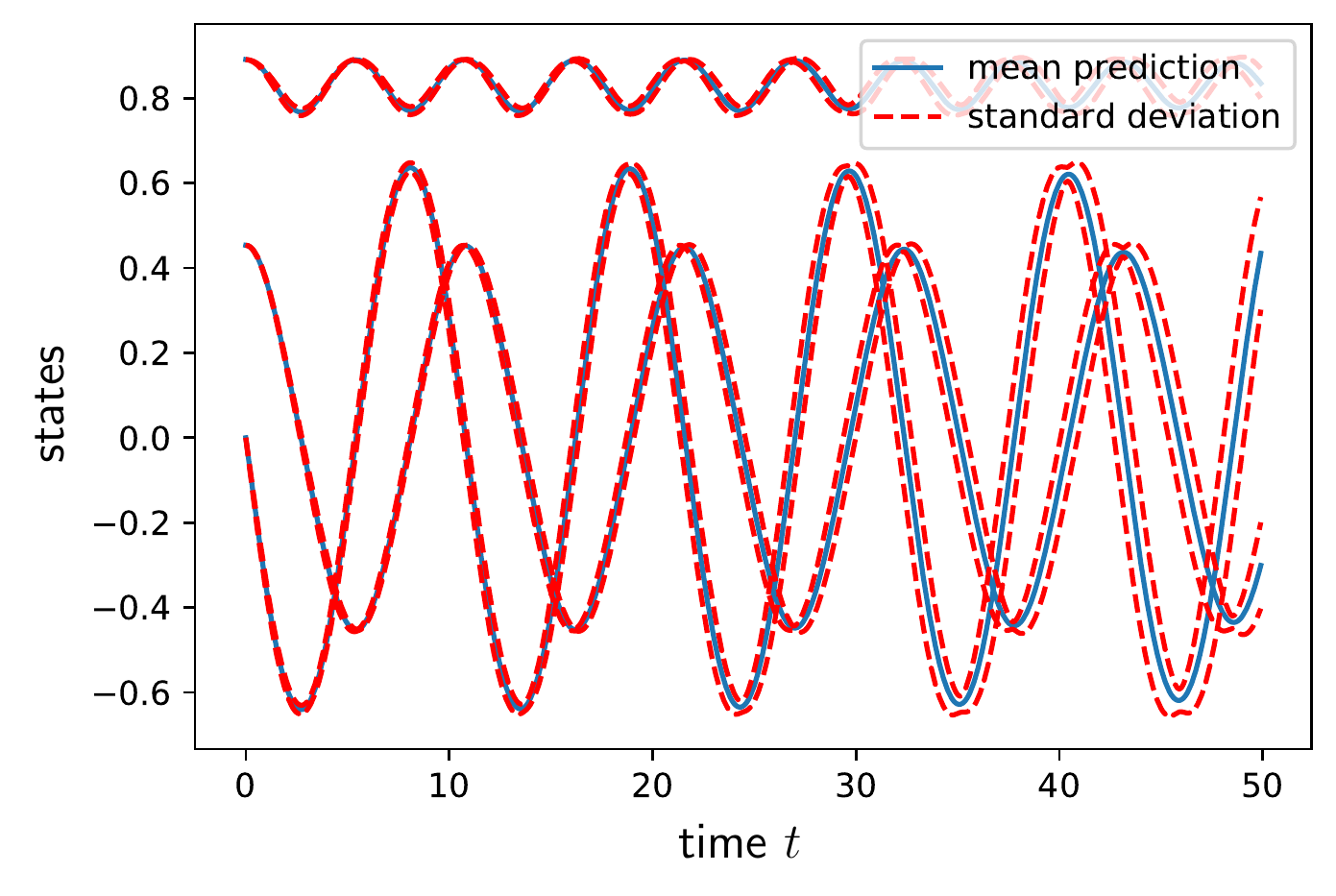}
		\caption{Rigid Body problem uncertainty}\label{subfig:varianceRB}
	\end{subfigure}
	
	\caption{Uncertainty information for the SGPD predictions from Section 6. Shown are mean (blue) and standard deviation (red) of 5 rollouts for the pendulum problem \ref{subfig:pendUncertainty}, the non-separable Hamiltonian \ref{subfig:ToyProblemUncertainty}, the two-body problem \ref{subfig:twoBodyUncertainty} and the rigid body dynamics \ref{subfig:varianceRB}.}
	\label{fig:var}
\end{figure*}

We add uncertainty information to the SGPD predictions by sampling $5$ trajectories $X_n^i, i =1,\dots, 5$ from the trained models (cf. Section 6). Mean $\mu_n$ and standard deviation $\sigma_n$ are approximated via
\begin{equation}
\begin{aligned}
\mu_n & = \frac{1}{5} \sum_{i=1}^{5} X_n^i \\
\sigma_n & = \sqrt{\frac{1}{4} \sum_{i=1}^{5} (X_n^i-\mu_n)^2}.
\end{aligned}
\end{equation}
Plots for all experiments are shown in Figure \ref{fig:var}.

\subsection{Rigid body dynamics} \label{ssec:RB}
\begin{figure*}[!h] 
\begin{subfigure}[h]{0.48\textwidth}
	\includegraphics[width=0.9\linewidth]{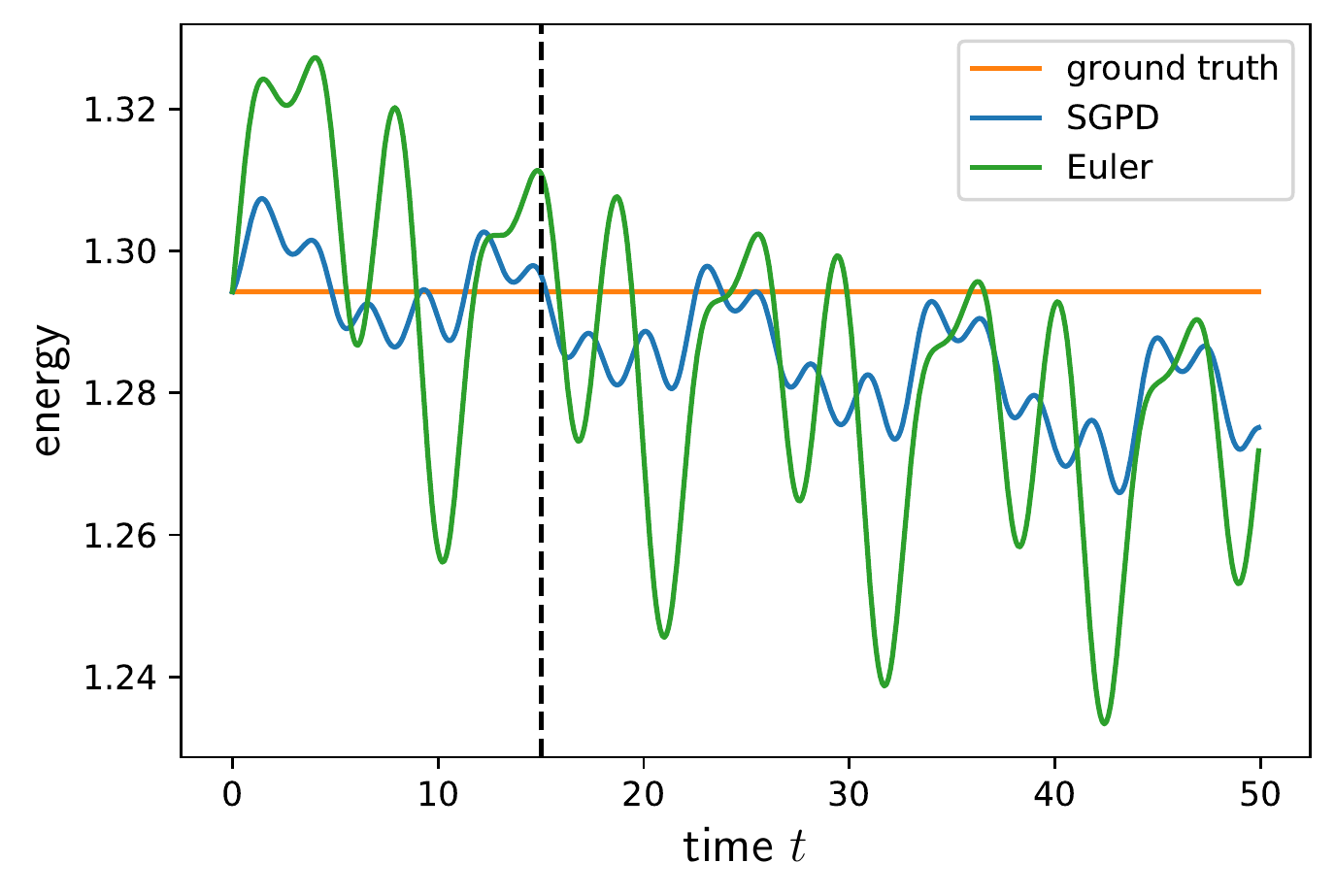} 
 \end{subfigure}
\begin{subtable}[h]{0.48\textwidth}
		\begin{tabular}{rcc}
	\noalign{\smallskip}  \hline \hline \noalign{\smallskip}
	method & energy err. & std. dev. \\
	\hline
	SGPD & $7\cdot 10^{-3}$   & $0.01$ \\
	Euler & $9\cdot 10^{-3}$  & 0.02  \\
	\noalign{\smallskip} \hline \noalign{\smallskip}
\end{tabular}
 \end{subtable}
\caption{Evaluation of the energy for the rigid body dynamics. Shown are the evolution of the energy (left) and the results for the analysis of the total energy (right). The SGPD preserves the energy more accuately than the explicit Euler method.}
\label{fig:energy}
\end{figure*}

We demonstrate that our SGPD method produces an accurate approximation to the total energy for the rigid body dynamics (\cf Section 6) in contrast to the standard explicit Euler dynamics. The constant ground truth energy $E$ is given $E = \frac{x_1^2}{2}+x_2^2+\frac{3 x_3^2}{2}$. We evaluate this equation for the SGPD and explicit Euler trajectories by averaging over the 5 samplied trajectories $E_n = \sum_{i=1}^5 E_n^i$. We compute the approximated average energy $\hat{E}=\sum_n \frac{E_n}{n}$ and evaluate the error $\|E-\hat{E} \|$ and the standard deviation $\sqrt{\sum_n \frac{|E_n-E|^2}{n-1}}$ along the trajectory. The results are shown in Figure \ref{fig:energy}. 
 
\subsection{Higher-order integrators} \label{ssec:higherOrder}
We demonstrate further benefits of our SGPD model learning scheme (cf. Section 4). In detail, we apply our method to a higher-order integrator. This yields a model that is less sensitive to varying step sizes for training and predictions than a model trained with the standard explicit Euler approach. 

\begin{figure*}[htb]
	\begin{subfigure}[t]{0.33\textwidth}
		\includegraphics[width=\textwidth]{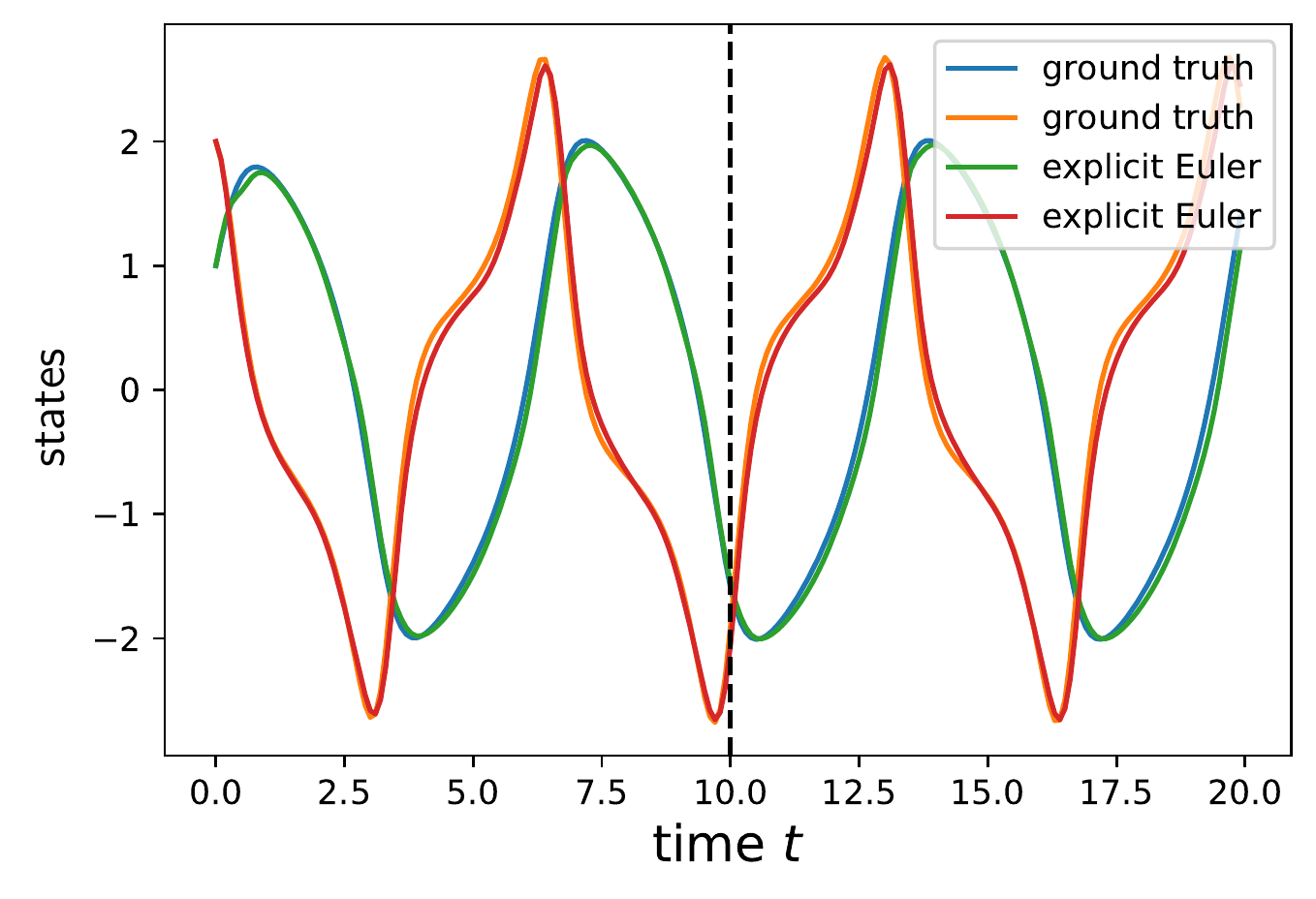}
				\caption{Euler, training step size}\label{subfig:explicit}
	\end{subfigure}
	\begin{subfigure}[t]{0.33\textwidth}
		\includegraphics[width=\textwidth]{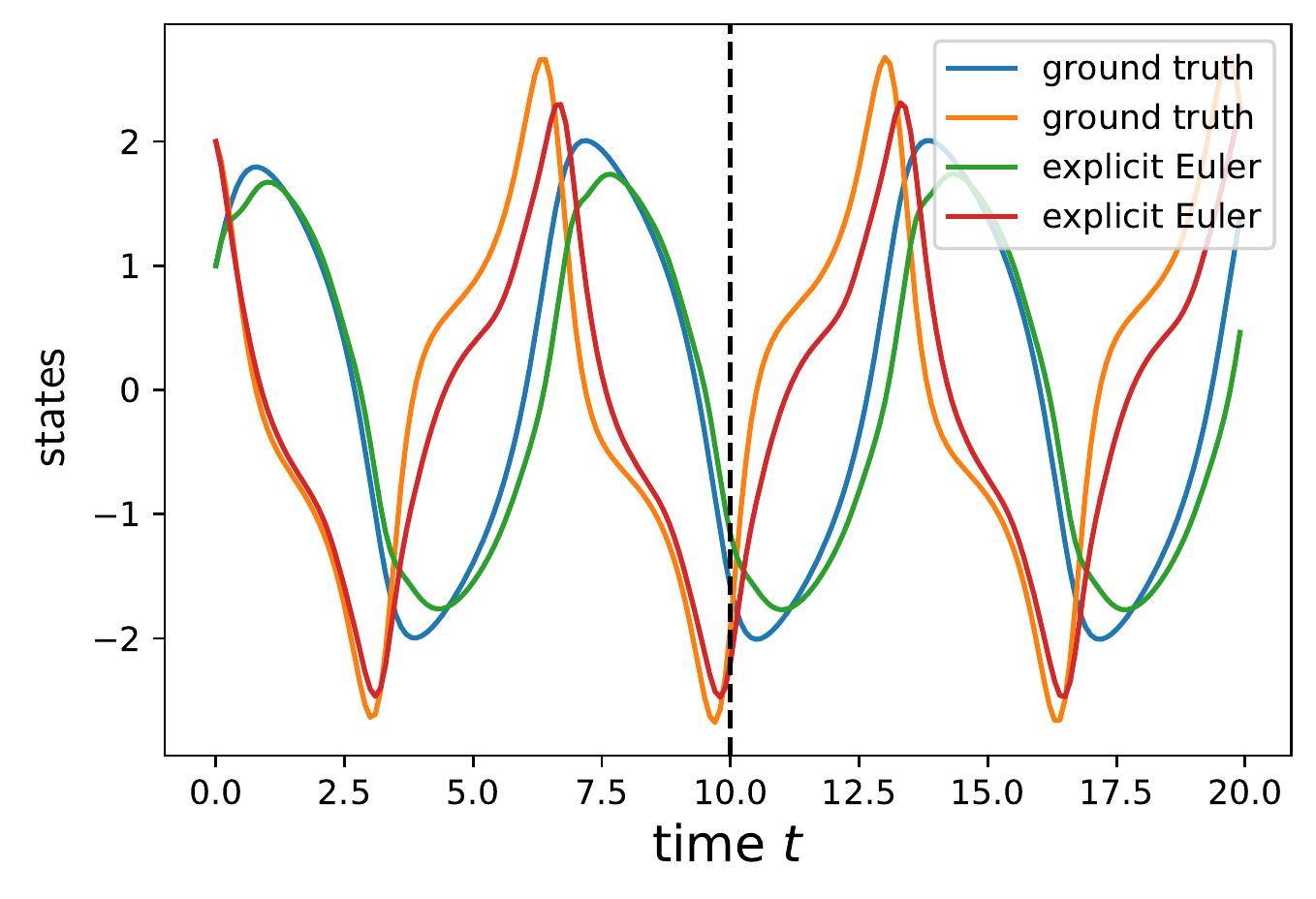}
				\caption{Euler, adaptive step size}\label{subfig:Adaexplicit}
	\end{subfigure}
	\begin{subfigure}[t]{0.33\textwidth}
		\includegraphics[width=\textwidth]{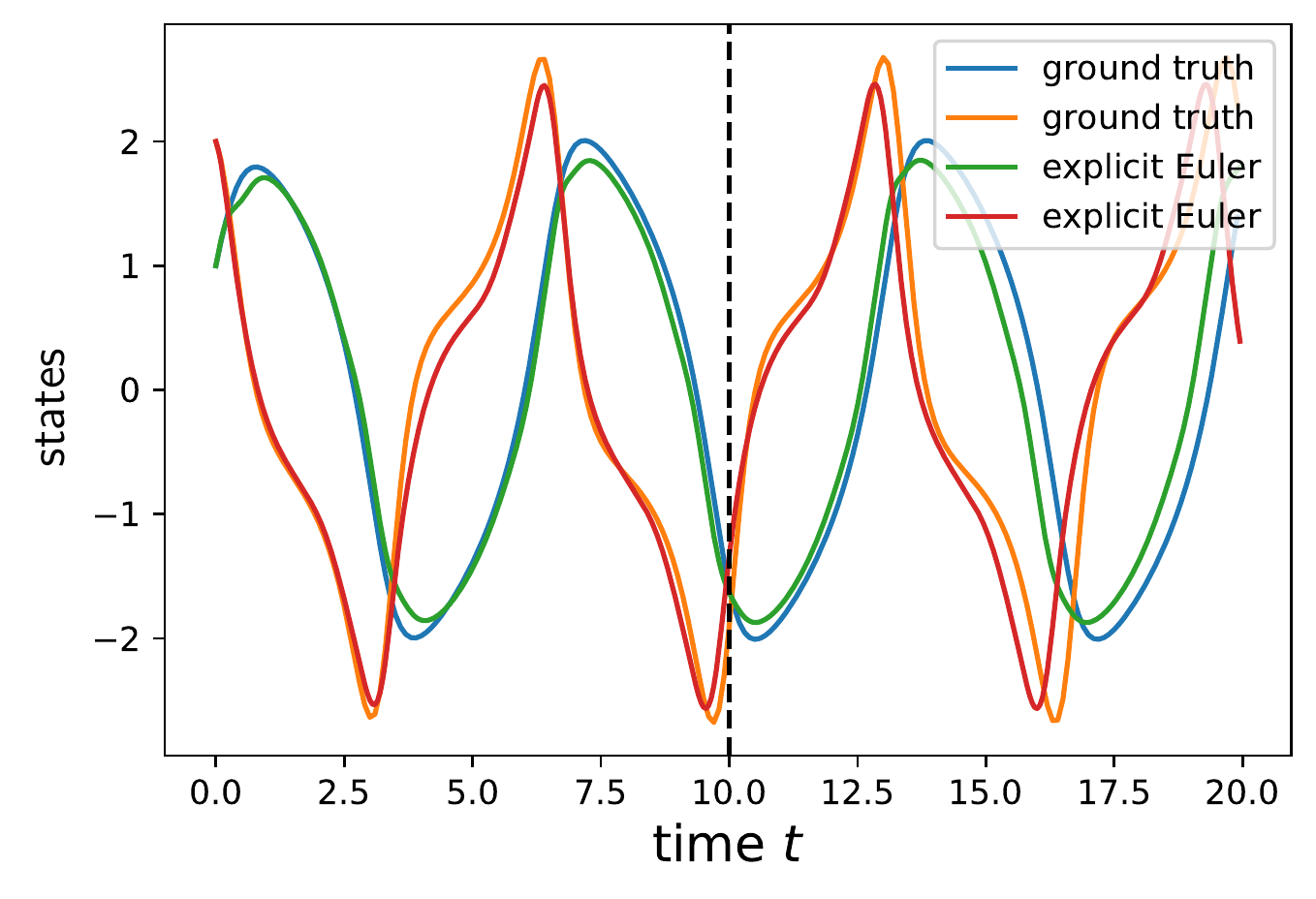}
				\caption{Euler, half step size}\label{subfig:Halfexplicit}
	\end{subfigure}
	\begin{subfigure}[t]{0.33\textwidth}
		\includegraphics[width=\textwidth]{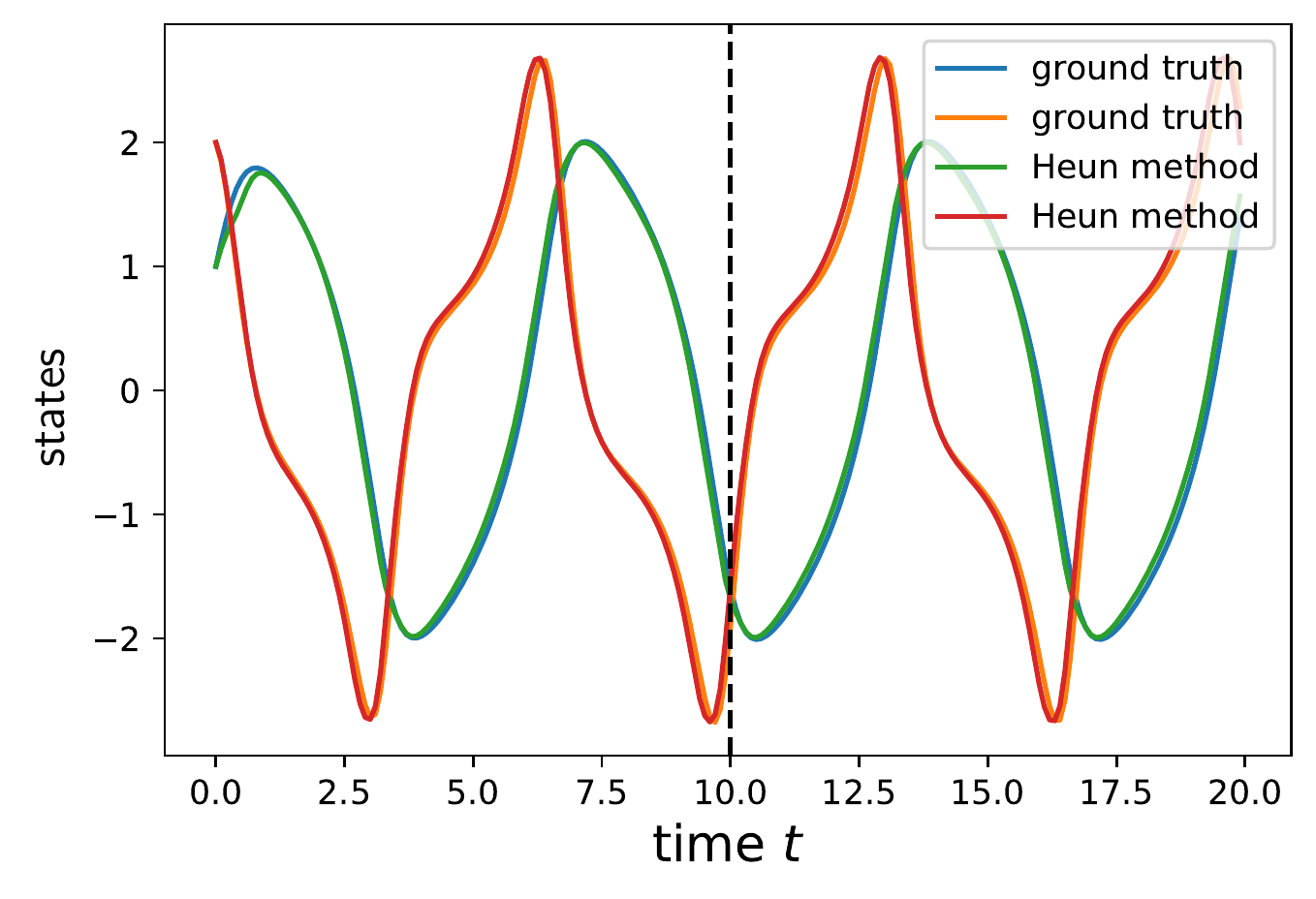}
		\caption{Heun, training step size}\label{subfig:midpoint}
	\end{subfigure}
	\begin{subfigure}[t]{0.33\textwidth}
		\includegraphics[width=\textwidth]{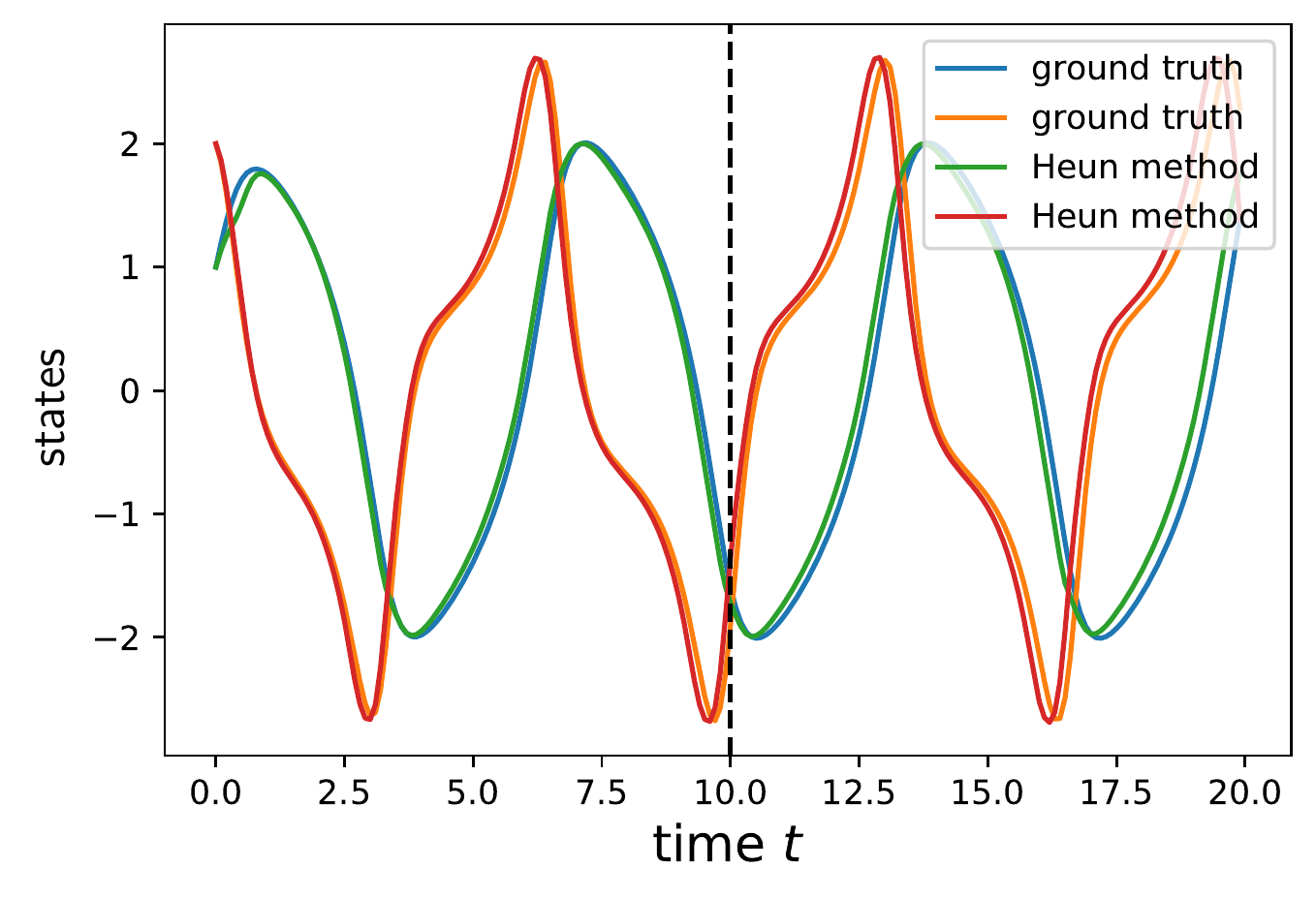}
		\caption{Heun, adaptive step size} \label{subfig:midpointAdaptive}
	\end{subfigure} 
	\begin{subfigure}[t]{0.33\textwidth}
		\includegraphics[width=\textwidth]{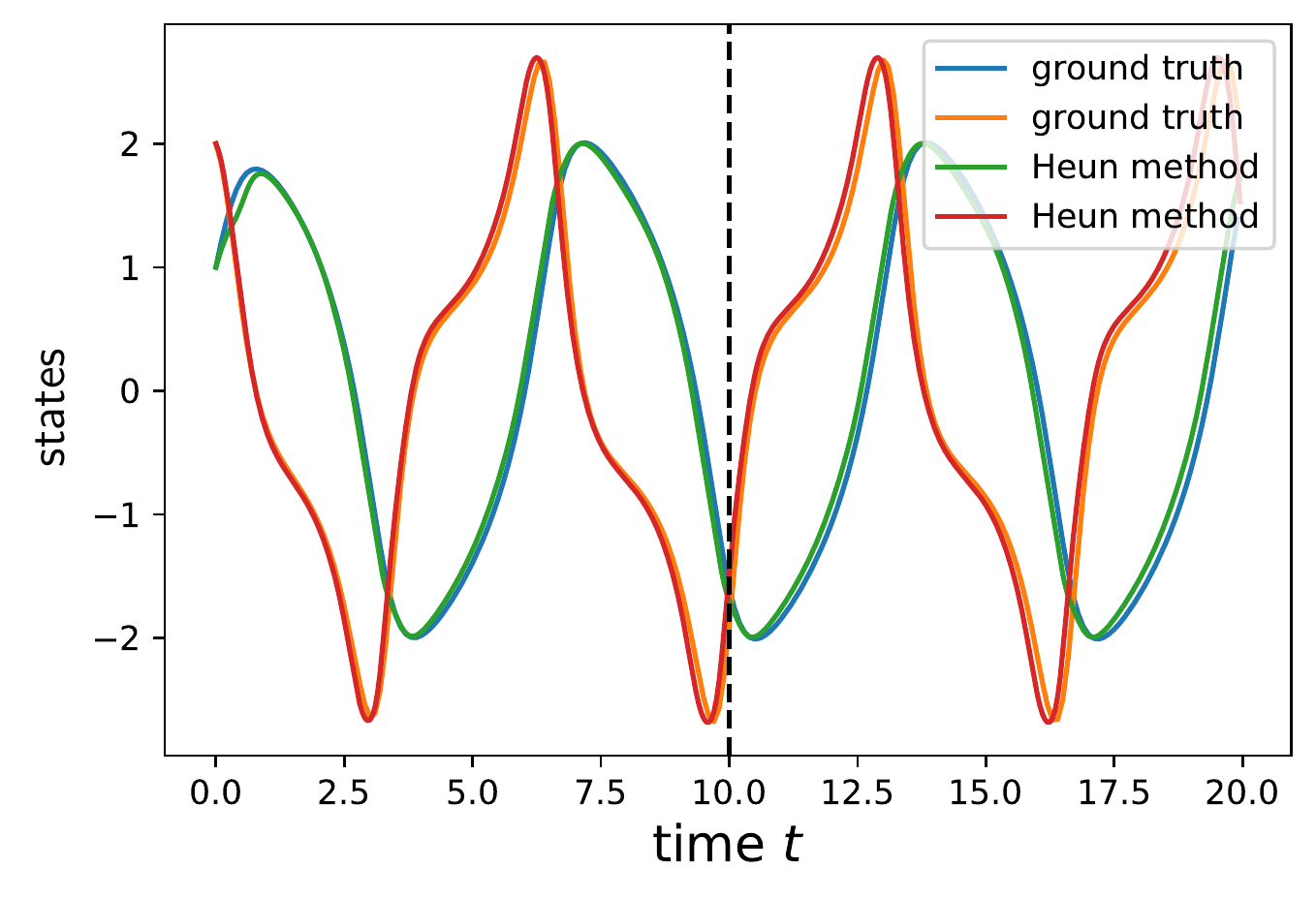}
		\caption{Heun, half step size}\label{subfig:midpoint005}
	\end{subfigure}
	\label{fig:midpoint}
		\caption{The Heun-based SGPD method is able to make accurate predictions with varying step sizes, while the explicit Euler method is adapted to the training step size $h=10^{-1}$. Shown are the results for the explicit Euler method (upper row) and the Heun-based SGPD method (lower row) for predictions with step size $h=10^{-1}$ (left), adaptive step size (middle) and step size $h = 5\cdot 10^{-3}$ (right).}
\end{figure*}

The number and choice of the internal stages determine the approximation order of a Runge-Kutta method $\psi_f$ \citep{hairerVol1} (cf. Section 3).
Let $x(t)$ be the ground truth solution of the continuous time dynamical system $\dot{x}(t)=f(x(t))$ and let $e = x(t)+h\psi_f(x(t))-x(t+h)$ be the local error.   
For a Runge-Kutta method with order $p$ it holds that
\begin{equation}\label{eq:error}
\Vert e \Vert_{\infty} \leq C h^{p+1}.
\end{equation} 
Intuitively, the higher the order, the smaller the error and the better the numerical approximation $x(n)$ (cf. Section 3). 
For the explicit Euler it holds that $p=1$. 
When choosing a higher-order Runge-Kutta integrator for dynamics model learning (cf. Section 3) the learned dynamics $\hat{f}$ provide a more accurate approximation to the ground truth dynamics $f$ than the standard explicit Euler dynamics model.
Thus, predictions can be made with varying step size.
We demonstrate the flexibility empirically by comparing the standard explicit Euler method to the second order Heun method 
\begin{equation}
\begin{aligned}
k_1 & = f(x_n) \textrm{, } k_2 = f(x_n+ h k_1) \\
x_{n+1} & = x_n+ \frac{h}{2} (k_1+k_2).
\end{aligned}
\end{equation}
Both models are trained with step size $h=10^{-1}$ on a horizon of 10 seconds. 
We demonstrate that our Heun-based SGPD method is able to perform predictions with varying step sizes in contrast to the explicit Euler method. 
To this end we compute predictions with half of the training step size $h=5\cdot 10^{-3}$. 
The SGPD method provides an accurate trajectory (Fig. \ref{subfig:midpoint005}), while the explicit Euler method fails (Fig. \ref{subfig:Halfexplicit}). 
Further, we demonstrate that the SGPD dynamics $\hat{f}$ are close to the ground truth dynamics $f$ by applying an adaptive step size solver to the predictions.  
This type of solver controls the numerical error $e$ in Eq. \eqref{eq:error} and adapts the step size accordingly. 
Again, the SGPD model provides accurate predictions (Fig. \ref{subfig:midpointAdaptive}), while the explicit Euler model fails (Fig. \ref{subfig:Adaexplicit}). 
Intuitively, the SGPD dynamics provide good results when beeing treated as continuous time dynamical system. 
The explicit Euler dynamics on the other hand correspond to a discrete-time model that is adapted to the training step size. 

\newpage
% !TEX root = supplement.tex

\section{Experimental setup} \label{sec:setup}
We give a more detailed explanation of our experimental setup. 
The experiments were performed on a desktop CPU.
The main libraries used for computation are \citet{NEURIPS2019_9015} and \citet{harris2020array}. For the plots and presentation \citet{Hunter:2007} was used. 
 
 \textbf{Initializiation: }
 We choose the ARD kernel  $k(x,x')=\sigma_f^2 \exp \left(-\sum_{i=1}^d \frac{(x_i-x_i')^2}{2 l_i^2}\right)$. 
 Hyperparameters include kernel parameters $\sigma_f$, $l$, variational sparse GP parameters  $\mu_{\xi,i}$,  $\Sigma_{\xi}$ (cf. Section 3), and the positions of the inducing inputs $\xi$.
 The initialization of the GP hyperparameters is shown in Table \ref{t:hyper} for the SGPD method and in Table \ref{t:hyperExp} for the explicit Euler method. 
 The number of inducing inputs per GP is denoted $N_{\xi}$. The total number of inducing inputs is always the same or higher for the explicit Euler model compared to the SGPD model. 
 For all experiments, except the two-body problem, all independent sparse GPs are initialized with the same hyperparameters. For the two-body problem the GP lengthscales are different for each dimension. Table \ref{t:l} contains the SGPD lengthscales and Table \ref{t:lEuler} the explicit Euler lengthscales for the two-body problem.

\begin{table}[htb]
	\caption{Hyperparameters SGPD}
	\centering
	\label{t:hyper}
	\begin{tabular}{rccccc}
		\noalign{\smallskip} \hline \hline \noalign{\smallskip}
		method \vline &  $\sigma_{f}^2$ & $l_i^2$ & $\mu_{\xi,i}$ & $\Sigma_{\xi}$ & $N_{\xi}$ \\
		\hline
		pendulum \vline &   0.01  & $\sqrt{2}$& $\mathcal{N}(0,0.05^2)$ & $10^{-8} \cdot \mathrm{I}$ & 9 \\
		two-body problem & $10^{-4}$ & Table \ref{t:l} &  $\mathcal{N}(0,0.05^2)$ & $10^{-8} \cdot \mathrm{I}$ & 20 \\ 
		non-separable system \vline &  $10^{-4}$ & $2$ & $\mathcal{N}(0,0.05^2)$ & $10^{-7} \cdot \mathrm{I}$ & 16 \\
		rigid body dynamics \vline & $10^{-3}$ & 1 & $\mathcal{N}(0,0.05^2)$ & $10^{-6} \cdot \mathrm{I}$& 9  \\		
		\noalign{\smallskip} \hline \noalign{\smallskip}
	\end{tabular}
\end{table}

 \begin{table}[htb]
 	\caption{Hyperparameters explicit Euler}
 	\centering
 	\label{t:hyperExp}
 	\begin{tabular}{rccccc}
 		\noalign{\smallskip} \hline \hline \noalign{\smallskip}
 		method \vline &  $\sigma_{f}^2$ & $l_i^2$ & $\mu_{\xi,i}$ & $\Sigma_{\xi}$ & $N_{\xi}$ \\
 		\hline
 		pendulum \vline &   0.01  & $\sqrt{2}$& $\mathcal{N}(0,0.05^2)$ & $10^{-8} \cdot \mathrm{I}$ & 9\\
 	    two-body problem & $10^{-4}$ & Table \ref{t:lEuler} &  $\mathcal{N}(0,0.05^2)$ & $10^{-8} \cdot \mathrm{I}$ & 20\\
 		non-separable system \vline &  $10^{-4}$ & $2$ & $\mathcal{N}(0,0.05^2)$ & $10^{-7} \cdot \mathrm{I}$ & 9 \\
 		rigid body \vline & $10^{-5}$ & 1 & $\mathcal{N}(0,0.05^2)$ & $10^{-8} \cdot \mathrm{I}$& 11 \\ 
 		\noalign{\smallskip} \hline \noalign{\smallskip}
 	\end{tabular}
 \end{table}
 
 \begin{table}[htb]
	\caption{Lengthscale SGPD two-body problem}
	\centering
	\label{t:l}
	\begin{tabular}{rcccc}
		\noalign{\smallskip} \hline \hline \noalign{\smallskip}
		dimension \vline &  $l_1^2$ & $l_2^2$ & $l_3^2$ & $l_4^2$  \\
		\hline
		$p_1$ \vline &   8.52  & 4.97 & 8.52 & 4.97\\
		$p_2$ \vline &  9.0 & 4.62 & 9.0 & 4.62 \\
		$p_3$ \vline &   8.52  & 4.97 & 8.52 & 4.97\\
		$p_4$ \vline &  9.0 & 4.62 & 9.0 & 4.62 \\
		$q_1$ \vline &  169 & 841 & 169 & 841 \\
		$q_2$ \vline & 256 & 324 & 129 & 324  \\
		$q_3$ \vline &  169 & 841 & 169 & 841 \\
	    $q_4$ \vline & 256 & 324 & 129 & 324  \\	
 		
		\noalign{\smallskip} \hline \noalign{\smallskip}
	\end{tabular}
\end{table} 

\begin{table}[htb]
	\caption{Lengthscale SGPD two-body problem}
	\centering
	\label{t:lEuler}
	\begin{tabular}{rcccccccc}
		\noalign{\smallskip} \hline \hline \noalign{\smallskip}
		dimension \vline &  $l_1^2$ & $l_2^2$ & $l_3^2$ & $l_4^2$ & $l_5^2$ & $l_6^2$ & $l_2^2$ & $l_8^2$ \\
		\hline
		$p_1$ \vline &   8.52  & 4.97 & 8.52 & 4.97 & 8.52  & 4.97 & 8.52 & 4.97  \\
		$p_2$ \vline &  9.0 & 4.62 & 9.0 & 4.62 & 9.0 & 4.62 & 9.0 & 4.62 \\
		$p_3$ \vline &   8.52  & 4.97 & 8.52 & 4.97 & 8.52  & 4.97 & 8.52 & 4.97\\
		$p_4$ \vline &  9.0 & 4.62 & 9.0 & 4.62 & 9.0 & 4.62 & 9.0 & 4.62 \\
		$q_1$ \vline &  169 & 841 & 169 & 841 & 169 & 841 & 169 & 841\\
		$q_2$ \vline & 256 & 324 & 129 & 324 & 256 & 324 & 129 & 324  \\
		$q_3$ \vline &  169 & 841 & 169 & 841 & 169 & 841 & 169 & 841 \\
		$q_4$ \vline & 256 & 324 & 129 & 324 & 256 & 324 & 129 & 324   \\	
		
		\noalign{\smallskip} \hline \noalign{\smallskip}
	\end{tabular}
\end{table} 
\textbf{Inducing inputs: }
For the pendulum problem with SGPD $\xi_p$ lie on a uniform grid on the interval $\left[-5,5\right]$ and $\xi_q$ lie on a a uniform grid on the interval $\left[-3,3\right]$. For the explicit Euler method $\xi$ lie on the corresponding two-dimensional interval with $-5\leq p\leq5$ and $-3 \leq q \leq 3$. 

For the two-body problem with SGPD it holds that $\xi_{p,1:4} \sim \mathcal{N}(-1.1,2.2^2), \xi_{q,1:4} \sim \mathcal{N}(-0.7,1.4^2)$. For the two-body problem with explicit Euler method the same configuration holds true on the corresponding 8-dimensional interval.  

For the non-separable Hamiltonian problem the inducing inputs are initialized on the uniform grid on the two-dimensional interval $\left[-0.5,0.5\right]$. 

For the rigid body dynamics with SGPD the inducing inputs are initialized on the two-dimensional interval with $x \in \left[-0.5,0.5\right]$ and $y \in \left[-0.7,0.7\right]$. For the explicit Euler method the inducing inputs are initialized via $x \in \mathcal{N}(-0.5,1), y \in \mathcal{N}(-0.7,1.7^2), z \in \mathcal{N}(0.7,0.2^2)$.  

\textbf{Dataset: }
For all experiments recurrent training (cf. Section 4) is performed on subtrajectories of the training trajectory. The trajectory is split into a dataset of all possible subtrajectories by choosing initial values via a sliding window with step size 1. Training is perfomed via epochs over the dataset of all subtrajectories. Shuffling and batching of the dataset is performed via Pytorch dataloader \citep{NEURIPS2019_9015}. Adam optimizer is applied for training. The training configuration is shown in Table \ref{t:training} for all systems. Our SGPD method and the standard explicit Euler method are trained with the same configurations. 

\begin{table}[htb]
	\caption{Tranining configuration}
	\centering
	\label{t:training}
	\begin{tabular}{rcc}
		\noalign{\smallskip} \hline \hline \noalign{\smallskip}
		system \vline & batch size & subtrajectory length \\
		\hline
		pendulum \vline &   1  & 10 \\
		two-body probem \vline &  5  & 50 \\
		non-separable system \vline & 1  & 10 \\ 
		rigid body  \vline & 1 & 20 \\
		\noalign{\smallskip} \hline \noalign{\smallskip}
	\end{tabular}
\end{table}

\textbf{Initial value: }
We observe noisy initial values $\hat{x}_0$ (cf. Section 1). During training we choose $x_0=\hat{x}_0$.

\textbf{Tolerance of the implicit solver: }
The implicit integrators require solving a nonlinear system of equations. We choose the Levenberg-Marquardt solver implemented in \citet{2020SciPy-NMeth} with the default configuration. 

\textbf{Loss function: }
The loss is computed via the ELBO (cf. Section 4). We balance the influence of the KL-divergence with respect to the total loss with factors $a$ and $b$. 
The loss $\mathcal{L}$ is then computed $\mathcal{L}=a \log p(\hat{x}_n|x_n)-b\mathrm{KL}(p(z)||q(z))$.
The chosen factors are documented in Table \ref{t:factors}. 
\begin{table}[htb]
	\caption{ELBO factors}
	\centering
	\label{t:factors}
	\begin{tabular}{rcc}
		\noalign{\smallskip} \hline \hline \noalign{\smallskip}
		system \vline & a & b  \\
		\hline
		pendulum \vline &   4  & $10^{-6}$ \\
		two-body probem \vline &  20 & $10^{-6}$ \\
		non-separable system \vline & 1 & $10^{-6}$  \\ 
		rigid body  \vline & 20 & 1  \\
		\noalign{\smallskip} \hline \noalign{\smallskip}
	\end{tabular}
\end{table}

\textbf{Decoupled sampling: }
We sample from GPs via decoupled sampling 
\begin{equation}
\begin{aligned}
f(x^{\star}|z, \xi)&=\underbrace{f(x^{\star})}_{\textrm{prior}}+\underbrace{k(x^{\star},\xi)k(\xi,\xi)^{-1}(z-f_z)}_{\textrm{update}} \\
& \approx \sum_{i=1}^{S} w_i \phi_i(x^{\star})+\sum_{j=1}^M v_j k(x^{\star},\xi_j)(z-f_z).
\end{aligned}
\end{equation}
For the ARD kernel $k(x,x')=\sigma_f^2 \exp \left(-\sum_{i=1}^d \frac{(x_i-x_i')^2}{2 l_i^2}\right)$ it holds that $\phi_i(x)=\sqrt{\frac{\sigma_f^2}{S}}(\cos x^T\omega_i,\sin x^T \omega_i)$. The parameters $\omega_i$ are sampled proportional to the spectral density of the kernel $\omega_i \sim \mathcal{N}(0,\Lambda^{-1})$, where $\Lambda = \textrm{diag}(l_1^2,l_2^2,\dots,l_d^2)$. For the ARD kernel this results in $2S$ feature maps $\phi(x) \in \mathbb{R}^{2S}$. Weights $w \in \mathbb{R}^{2S}$ are sampled from a standard normal $w_i \sim \mathcal{N}(0,1)$ \citep{hegde2021bayesian}. We choose $S = 10000$ for our experiments. 

\textbf{Model selection: }
Let $\hat{X}_i=\frac{1}{5} \sum_{j=1}^5 \hat{X}_i^j, i = 1 \dots n$ be the average of 5 rollouts and $\hat{x}_{1:n}$ the noisy observations.
in order to obtain the best model we compute $e = \sum_{i=1}^n \Vert \hat{X}_n-\hat{x}_n \Vert^2$ at the beginning of every epoch. After training we choose the model with the smallest error $e$. 
In case the learning rate is adapted during training we apply the model selection criterion to models corresponding to the smallest learning rate. 

\textbf{Learning rates and model selection: }
The pendulum problem is trained for 149 epochs with an initial learning rate of $10^{-2}$ for the SGPD and the explicit Euler method.
Model selection is performed over all epochs.

The two-body problem is trained for 149 epochs with an initial learning rate $10^{-2}$. After 100 epochs the learning rate is lowered to $10^{-3}$ for the SGPD and the explicit Euler method. 
Model selection is performed on the final 45 epochs for the SGPD and the explicit Euler method. 

For the non-separable Hamiltonian system our SGPD method is trained for 10 epochs with an initial learning rate of $10^{-4}$. After 2 epochs the learning rate is set $10^{-2}$, after 5 epochs it is set $10^{-5}$.
Model selection is performed on the final 5 epochs. 

The nonseparable Hamiltonian system with explicit Euler method is trained for 49 epochs with learning rate of $10^{-3}$. 
Model selection is performed over all epochs. 

The rigid-body dynamics with SGPD is trained for 11 epochs with an initial learning rate of $10^{-2}$. 
Each two epochs the learning rate is scaled with a factor $10^{-1}$ up to the final learning rate $10^{-5}$. 
Model selection is performed over the final 4 epochs. 

The rigid body dynamics with explicit Euler method is trained for 20 epochs with an initial learning rate of $10^{-2}$. 
After 10 epochs the learning rate is set $10^{-3}$, after 10 epochs it is set $10^{-4}$. 
Model selection is performed over all epochs.

\textbf{Training data: } Training data are generated from ground truth and disturbed with noise as described in Section 6.
Initial values are chosen $(2,2)$ for the pendulum problem, $(0,-1.5/4)$ for the non-separable Hamiltonian system and $(\cos(1.1), 0, \sin(1.1))$ for the rigid body dynamics and $(1.144,0.880,-0.241,0.313,-1.144,-0.880,0.241,-0.313)$ for the two-body problem.  For the two-body problem we generate the data via the generator provided in \citet{NIPS20199672} with \texttt{orbit, settings = get\_orbit(state, t\_points=201, t\_span = [0, 30], rtol = 1e-10)} and numpy random seed 30. 
\newpage
\bibliography{bib}